\documentclass[draft.pdflatex,sn-apa,iicol]{sn-jnl}
\usepackage[T1]{fontenc}

\usepackage{graphicx}\usepackage{multirow}\usepackage{amsmath,amssymb,amsfonts}\usepackage{amsthm}\usepackage{mathrsfs}\usepackage[title]{appendix}\usepackage{xcolor}\usepackage{textcomp}\usepackage{manyfoot}\usepackage{booktabs}

\theoremstyle{thmstyleone}

\theoremstyle{thmstyletwo}

\theoremstyle{thmstylethree}\newtheorem{definition}{Definition}

\usepackage[]{natbib}

\newcommand{\todo}[1]    {\par\begin{center}\fbox{
			\begin{minipage}[t]{0.95\columnwidth}
				{\bf TODO: }\\ #1
			\end{minipage}
}\end{center}}

\usepackage{epsfig}
\usepackage{color}
\usepackage{epstopdf}
\usepackage{comment}
\usepackage{multirow}

\usepackage[active]{srcltx}

\usepackage{tikz}
\usetikzlibrary{calc,angles,quotes}

\usetikzlibrary{arrows.meta}
\usetikzlibrary{decorations.pathreplacing}
\usepackage{cleveref} \usepackage{subcaption} \usepackage{colortbl}

\usepackage{siunitx}  \newcommand{\m}{\unit{\metre}}

\newcommand{\ms}{\unit{\meter/\second}}

\crefname{equation}{Eq.}{Eqs.}
\crefname{figure}{Fig.}{Figs.}
\crefname{appendix}{Appendix}{Appendices}
\crefname{item}{Step}{Steps}
\crefname{section}{Sec.}{Secs.}
\crefname{table}{Tab.}{Tabs.}
\crefname{definition}{Def.}{Defs.}

\newcommand{\Lt}[1]{\vec{L_{ij}^{#1}}}
\newcommand{\Rt}[1]{\vec{R_{ij}^{#1}}}
\newcommand{\Ct}[1]{\vec{C_{ij}^{#1}}}

\begin{document}

\title[Fault-Tolerant Collective Motion]{Bugs with Features: Vision-Based Fault-Tolerant Collective Motion Inspired by Nature}

\author*[1]{\fnm{Peleg} \sur{Shefi}}\email{shfpeleg@gmail.com}

\author[2,3]{\fnm{Amir} \sur{Ayali}}\email{ayali@tauex.tau.ac.il}

\author[4,1]{\fnm{Gal} \sur{A. Kaminka}}\email{galk@cs.biu.ac.il}

\affil*[1]{\orgdiv{Computer Science Department}, \orgname{Bar Ilan University}, \orgaddress{\city{Ramat Gan}, \postcode{5290002},  \country{Israel}}}

\affil[2]{\orgdiv{School of Zoology}, \orgname{Tel Aviv University}, \orgaddress{\city{Tel Aviv}, \postcode{6997801}, \country{Israel}}}

\affil[3]{\orgdiv{Sagol school of Neuroscience}, \orgname{Tel Aviv University}, \orgaddress{\city{Tel Aviv}, \postcode{6997801}, \country{Israel}}}

\affil[4]{\orgdiv{Brain Research Center}, \orgname{Bar Ilan University}, \orgaddress{\city{Ramat Gan}, \postcode{5290002}, \country{Israel}}}

\abstract{In \textit{collective motion}, perceptually-limited individuals move in an ordered manner, without centralized control. The perception of each individual is highly localized, as is its ability to interact with others. While natural collective motion is robust, most artificial swarms are \emph{brittle}.
This particularly occurs when vision is used as the sensing modality, due to ambiguities and information-loss inherent in visual perception.
This paper presents mechanisms for robust collective motion inspired by studies of locusts.
First, we develop a robust distance estimation method that combines visually perceived horizontal and vertical sizes of neighbors.
Second, we introduce \textit{intermittent locomotion} as a mechanism that allows robots to reliably
detect peers that fail to keep up, and disrupt the motion of the swarm. We show how
such faulty robots can be avoided in a manner that is robust to errors in classifying them as faulty.
Through extensive physics-based simulation experiments, we show dramatic improvements to swarm resilience when using these techniques. We show these are relevant to both distance-based \emph{Avoid-Attract} models, as well as to models relying on \emph{Alignment}, in a wide range of
experiment settings.
 }


\maketitle
\sloppy

\section{Introduction}
\label{ch:Introduction}
Ordered collective motion (\emph{flocking}) is a swarm phenomenon in which individuals---limited to highly localized interactions---move in a globally ordered fashion, without centralized control.
Despite a large variety in sensing, actuation, and cognitive capabilities, collective motion is widespread in nature, from insects, birds, and fish, to mammals and humans~\citep{emlen52,katz11,giardina08,ariel15,ginelli15,wolff73,sayin25,calovi18}.

Many collective motion models, intended to model this natural phenomenon, have also been extended and utilized to synthesize collective motion in virtual or physical robots~\citep{huth92,helbing95,vicsek95,reynolds87,mataric97,ferrante12,romanczuk12,strombom11,bastien20,mezey25,krongauz24,cmot10,tist18,qi23,moshtag05,moshtag09}.
An  important motivation for synthetic swarms is the observed robustness of natural swarms, both to perceptual processes that are inherently fallible, as well as to catastrophic member failures (e.g., individuals that cannot keep up).

Unfortunately, existing robot swarm models are generally not nearly as robust as their natural inspirations,
particularly when vision is used as the primary sensing modality.
As we show (and discussed in~\cref{ch:Related Work and Background}), inherent ambiguities in distance perception, and information-loss due to occlusions can cause degradation to the ordered motion of the swarm, even when each robot is behaving nominally~\citep{moshtag09,krongauz24}.
Difficulty in visually distinguishing ego-motion and the motion of others in the visual signal leads to an
inability to detect (and avoid) individuals that are unable to keep up.
Even a small proportion of such robots can completely disrupt collective order~\citep{winfield06,bjerknes13,shefi24}.

This paper seeks to improve the robustness of vision-based collective motion.
We focus on the perception processes that extract the information needed for the computation of the desired motion of the individual.  We remove the assumption of idealized sensing, and instead examine limitations imposed by visual sensors: occlusions, inability to sense range, perception of movement that combines multiple sources of motion, and more.

Inspired by recent studies of the visual perception of individual locust~\citep{bleichman24}, we present a method for robust estimation of distance from vision, combining the perceived vertical and horizontal sizes of  robots (neighbors) perceived by each individual (focal robot).
The method corrects for the errors in the estimated distance, which may be caused by the orientation of the neighbors (relative to the focal observer).
Such errors are marginal (or non-existent) for cylindrical robots, but in elongated (rectangular box) robots, these errors have a devastating effect on the order, as we show.

Then, we show that the use of intermittent locomotion---likewise found in locusts~\citep{ariel14}---allows nominal (healthy) robots to identify faulty peers that are too slow to keep up. Each robot asynchronously switches between moving and pausing states.
When paused, the robot observes others and verifies that their movement is nominal. When moving, it avoids faulty neighbors, so they do not disrupt its decisions.

Intermittent \emph{Pause} and \emph{Go} locomotion  (P\&G) is deceptively simple. Pausing robots may be mistakenly categorized by others as faulty, so pause intervals should be kept short.
However, long stretches of locomotion
prevent the robot from detecting failures in others.
We developed mechanisms that use a stochastic approach for interacting with detected faulty neighbors (that may have been misclassified as such).
We show that these mechanisms mitigate the negative effects of faulty robots
 on the ordered motion of the swarm.

Both the pause-and-go strategy and the robust distance estimation are used to allow a visual sensor to be used with an \textit{Avoid-Attract} (AA) model~\citep{ferrante12,qi23}.
This model relies only on the distance and bearing to neighbors.
We use the physics-based ARGoS3 simulator~\citep{pinciroli12} to empirically evaluate collective motion methods in swarms of simulated \textit{Nymbot} robots~\citep{frontiers23amir}.
These are rectangular, two-wheeled differential-drive robots, designed to replicate the size, maximum speed,
and turning dynamics of locust nymphs.

The experiments validate the effectiveness of the methods, demonstrating their potential to improve the fault tolerance of the collective motion. When the AA model is used with the robust distance estimator, the resulting \emph{AA-V} model is able to maintain order despite the limitations of visual sensing. Building on this model to add intermittent locomotion, the \emph{AAPG-V} model is able to maintain order in face of severe faults.

The paper is organized as follows. ~\cref{ch:Related Work and Background} surveys previous work on robust and vision-based collective motion.
\cref{sec:baseline} presents the baseline Avoid-Attract model and preliminaries.
\cref{ch:pereception_challenges} introduces visual sensing and its impact on robust estimation of distance to neighbors.
\cref{ch:problem_analysis} presents intermittent locomotion and its application in identifying faulty robots.
\cref{ch:Discussion} then discusses how the techniques can be used with other collective motion models, and in cases where faulty robots slow down rather than stop.
\cref{ch:Summary} concludes the paper.
Technical appendices provide additional details.

 \section{Background}
\label{ch:Related Work and Background}
Coordinated collective motion is an area of research that spans multiple disciplines.
Our focus is on models that facilitate the generation of collective motion in agents. Numerous such models exist, e.g.,~\citet{reynolds87,huth92,vicsek95,helbing95,mataric97,moshtag05,moshtag09,cmot10,strombom11,vicsek2012collective,ferrante12,romanczuk12,ariel15,tist18,bastien20,joshi22,qi23,castro24,krongauz24,mezey25,sayin25}. While models vary in their specification of perceptual, computational, and actuation processes, they generally agree that collective motion results from repeated, decentralized decisions of individuals responding to their locally-sensed neighbors.

\citet{fine13} argue that most models ignore the impact of sensing and perceptual processes on the generated collective motion. They offer a unifying perspective that partition the individual member control process into five distinct stages that work in series, as a pipeline:
\textit{sensing}, \textit{neighbor detection}, \textit{neighbor selection}, \textit{motion computation}, and \emph{physical motion}.
Errors or commitments made in earlier stages greatly affect later stages.

\paragraph{Sensing and Detecting Neighbors}
Recent investigations have been exploring the use of visual perception for collective motion.
As a result, key techniques have begun to emerge for translating visual imagery into a set of detected neighbors, each associated with a set of geometric features used in later stages.
Some previous investigations rely on both distance and orientation (sometimes by different modalities)~\citep{moshtag05,moshtag09,zhou24,ito24,krongauz24}.
Others rely on range and bearing~\citep{ferrante12,qi23,bastien20,mezey25}, or relative velocity~\citep{romanczuk12}.

The estimation processes are different, depending on the feature. Bearing (relative angle to the neighbor) is deduced directly from the pixel position in images~\citep{bastien20,romanczuk12,castro24,mezey25}. Range may be estimated from the size of the horizontal angles~\citep{krongauz24,bastien20,mezey25} or vertical angles~\citep{moshtag05,moshtag09,ito24,harpaz21,zhou24}  subtended on the neighbor  (the angular width and height of the perceived neighbor, respectively).
Inspired by recent findings in locusts~\citep{bleichman24}, we show that combining information from both vertical and horizontal angles yields a much better estimator than either alone.

\paragraph{Selecting Nominal Neighbors}
Most collective motion models treat all detected robots as nominal (healthy). However, various types of failures in individual robots can lead to catastrophic effects on collective behavior~\citep{bjerknes13,christensen09,tarapore17,winfield06,carminati24}. One of the most severe failures---the focus of our work---is when a robot stops functioning permanently, while its nominal neighbors do not recognize it as faulty~\citep{bjerknes13,tarapore17,winfield06}.
For instance, if faulty robots slow below nominal thresholds (or even stop), their nominal neighbors may become dynamically ``anchored'' to them.
 This propagates through the swarm by local interactions, preventing the swarm from advancing.
There are several approaches to handling such failures.

\subparagraph{Active fault detection}
Active methods rely on robots generating signals or monitoring processes for fault detection.
They can be very effective, but rely on the reliability of the signal, and its reception.
\textit{Endogenous} methods rely on self-monitoring by the robots, and signal if a fault is recognized~\citep{christensen08,dixon01,zhuo05,millard14}.
\textit{Active signaling} methods use periodic \emph{heartbeats} (e.g., radio or light), with faults inferred when signals stop~\citep{christensen09,tarapore17}.
However, if robots continue broadcasting despite immobility, faults may be undetected.
\emph{Mutual monitoring} methods have robots cross-monitor each other consistency~\citep{aamas11eli,icra10anomaly,millard14}.
These are extended to swarms using consensus or behavioral feature comparisons (relying on prediction models)~\citep{tarapore15,tarapore17,oKeeffe18,carminati24}.

\subparagraph{Passive fault detection}
A different set of approaches utilizes fault detection methods that do not require active participation by neighbors.
\citet{jair00} propose to simulate others' internal processes (using plan recognition) to detect disagreements between robots executing coordinated plans. Putting aside the computational requirements~\citep{dars02brett}, the method assumes prior knowledge of all plans and the ability to visually recognize the motions of others. Similarly,
\citet{millard14} and \citet{millard14b}  propose having robots simulate their peers’ controllers using onboard models, focusing on predicting neighbors’ future spatial states. Practically, deviations of observed motions from these predictions signal a fault. However, as such discrepancies naturally occur and accumulate over time (especially in a noisy environment with collisions), this requires
periodic re-initialization via communication or visual information.

However, reliance on vision is at the heart of the challenge to detect robots that fail to keep up.
Naturally, if the speed of a neighbor is perceived to be below par, the robot can be marked as faulty, and the focal robot can treat it as such.

Unfortunately, to estimate the speed of a neighbor, the ego-motion of the observer has to be accounted for in the perceptual input.  As a nominal robot moves, its visual sensors capture relative movement of neighbors, combining the visual effects of its own ego-motion with those resulting from the motions of its neighbors. Isolating the different sources of movement to estimate the velocity of the neighbor is infeasible if the observer does not know its own actual movement.

To address this issue, we investigate a \textit{behavioral} approach, whereby nominal robots pause and go intermittently, so as to make it easy for themselves to visually perceive the motion of others. During pauses, ego-motion is nullified, and so any perceived motion is due to neighbors. The neighbors' velocities can then be measured sufficiently accurately to detect faulty members. They are then avoided during the movement phase.

\section{A Baseline Model and Experimental Setup}
\label{sec:baseline}
We begin by describing (\cref{sec:aa_model}) an established baseline model that will serve as the scaffold
for the investigation of robust vision-based collective motion. We then describe the experimental environment that we use to evaluate the techniques presented in the rest of the paper (\cref{sec:experimental_env}).

\subsection{Theoretical Collective Motion: The Avoid-Attract (AA) Model}
\label{sec:aa_model}
We consider a collective of \textit{rectangular box} robots. All the robots are mechanically identical (have same width, length, height). The set of robots is denoted $\mathcal{A}$.
Individual robots have only local knowledge: neither the set $\mathcal{A}$ nor their position in a global coordinate system is known to them.
We refer to a robot observing those around it as the \emph{focal robot}, and those within its local sensing range as \emph{neighbors}.

We adopt a range-and-bearing interaction model, thereby reducing reliance on ego-motion and orientation estimates required by many classical approaches, starting with~\citet{vicsek95}. Instead, we employ the \emph{Avoid--Attract} (AA)  model~\citep{ferrante12}, in which robots seek to maintain a preferred inter-robot distance. AA is both simple and effective.

We briefly describe the model below, using the unifying framework presented by~\citet{fine13}.
Appendix~\ref{app:aa} provides a full description and nomenclature of the symbols used.
In particular, \cref{tab:aa_params_appendix} lists the parameters and default values used.

\subparagraph{Sensing} The AA model makes the common assumptions of an \textit{oracle sensor} that provides all needed information with perfect accuracy and zero computational cost. Each robot is equipped with an idealized (error-free, never-occluded) sensing mechanism, characterized by a sensing limit radius
\( r_{\text{sense}} \), and omni-directional (\ang{360}) field of view. This allows the focal robot to measure range (distance) and bearing angle (relative to the focal robot's heading) to any and all robots within the sensing radius, regardless of occlusions.
The dimensions of the robots do not impact the sensed information.

\subparagraph{Neighbor detection}
\citet{ferrante12} uses the idealized range and bearing sensor to detect all robots whose \emph{center} position lies within the sensing radius. Each neighbor $j$ is represented by the vector
\begin{equation}
\Ct{} = (r_{ij}, \angle \beta_{ij}),
\label{eq:cvector}
\end{equation}
given in the polar coordinate system centered at the focal robot center, and aligned with its heading.
Here, $\beta_{ij}$ is the relative bearing
angle between the focal robot’s $x$-axis (aligned with its body) and the line connecting the two robot
centers. $r_{ij}$ is the distance between the centers along this line.
When clear from the context, we refer to these also as $r_C, \beta_C$.

A focal robot $i\in \mathcal{A}$ detects a neighbor $j\in \mathcal{A}$ if \( r_{ij} \leq r_{\text{sense}} \).
The set of all detected neighbors of robot $i$ at time $t$ is denoted $\mathcal{D}_i^t\subseteq \mathcal{A}$. Given the idealized nature of the sensor, no robot within $r_{sense}$ is ever occluded.

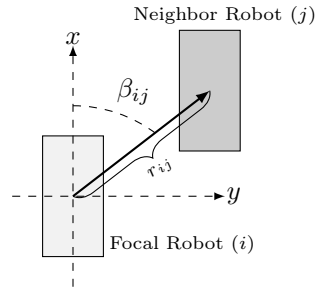
\begin{figure}[htbp]
	\centering
	\begin{tikzpicture}[scale=0.8]
\filldraw[fill=gray!10] (-0.5,1) rectangle (0.5,-1);
            \node at (1.8,-0.8) {\footnotesize Focal Robot ($i$)};

\filldraw[fill=gray!40] (1.75,0.75) rectangle (2.75, 2.75);
            \node at (2.5,3.0) {\footnotesize Neighbor Robot ($j$)};

\draw[-latex, thick] (0,0) -- (2.25,1.75);

\draw[decorate,decoration={brace,amplitude=7pt,mirror}]
  (0,0) -- (2.25,1.75)
  node[midway,sloped,below=6pt] {\footnotesize $r_{ij}$};

\draw[-latex,dashed] (0,-1.5) -- (0.0,2.5) node[above, fill=white, inner sep=0.5pt] {$x$};;
            \draw[-latex,dashed] (-1.0,0.0) -- (2.5,0.0) node[right, fill=white, inner sep=0.5pt] {$y$};;

\node[inner sep=1pt] at (60:2cm) {$\beta_{ij}$};
            \draw[dashed] (90:1.5cm)
              arc[start angle=90, end angle={-atan(1.75/2.25)+90}, radius=2.25cm];

	\end{tikzpicture}
	\caption{Focal robot’s coordinate system and relative neighbor location. The \textit{dashed arc} marks the bearing angle $\beta_{ij}$. The \textit{thick solid black arrow} measures the distance $r_{ij}$. The vertical $x$ axis marks the focal robot's current heading. }
	\label{fig:robot_rb}
\end{figure}

\cref{fig:robot_rb} shows a focal robot $i$ and a detected neighbor robot $j$, along with the associated polar measurements.

\subparagraph{Neighbor selection}
The set of selected neighbors for robot $i$ at time $t$ is denoted $\mathcal{N}_i^t$. Since the AA model does not distinguish faulty and nominal robots, it selects the set of all detected neighbors $\mathcal{D}_i^t$, i.e., $\mathcal{N}_i^t=\mathcal{D}_i^t$.

\subparagraph{Motion computation and control} Every robot independently seeks to maintain a distance $r_d$ from all its neighbors ($r_d$ is the same for all robots).  Each robot $i\in \mathcal{A}$ computes its intended motion vector at time $t$ ($\vec{F_i^t}$), by aggregating avoid/attract forces
with its neighbors $\mathcal{N}_i^t$~\citep{ferrante12}:

\begin{equation}
    \vec{F_i^t} := \sum_{j\in \mathcal{N}_i^t} \vec{f_{ij}^t}
    \label{eq:avoid-attract}
\end{equation}
where \citet{qi23} defines $\vec{f_{ij}}$ using gain $K_f\geq0$:
\begin{equation}
 \vec{f_{ij}^t}:=(K_f\cdot\frac{r_{ij} - r_d}{r_{ij}^2}, \angle\beta_{ij}).
 \label{eq:rjvector}
\end{equation}

A positive fraction results in attraction (too far), and a negative fraction leads to avoidance (too close).
This flocking vector is converted into linear velocity (\( u \)) and angular velocity (\( \omega \)) using magnitude-dependent motion control (MDMC)~\citep{ferrante12}. The full AA model and control equations are detailed in Appendix~\ref{app:aa}.  In our experiments, we use $K_f=1$.

\subsection{Experimental Settings}
\label{sec:experimental_env}
We use the ARGoS~3 platform~\citep{pinciroli12}, a physics-based robot simulator, to simulate the \textit{Nymbots}, two-wheeled differential-drive robots
designed to mimic locust nymphs in size, speed, and turning rate, enabling biologically grounded swarm simulations.

\begin{figure}
	\centering
\includegraphics[width=0.6\columnwidth]{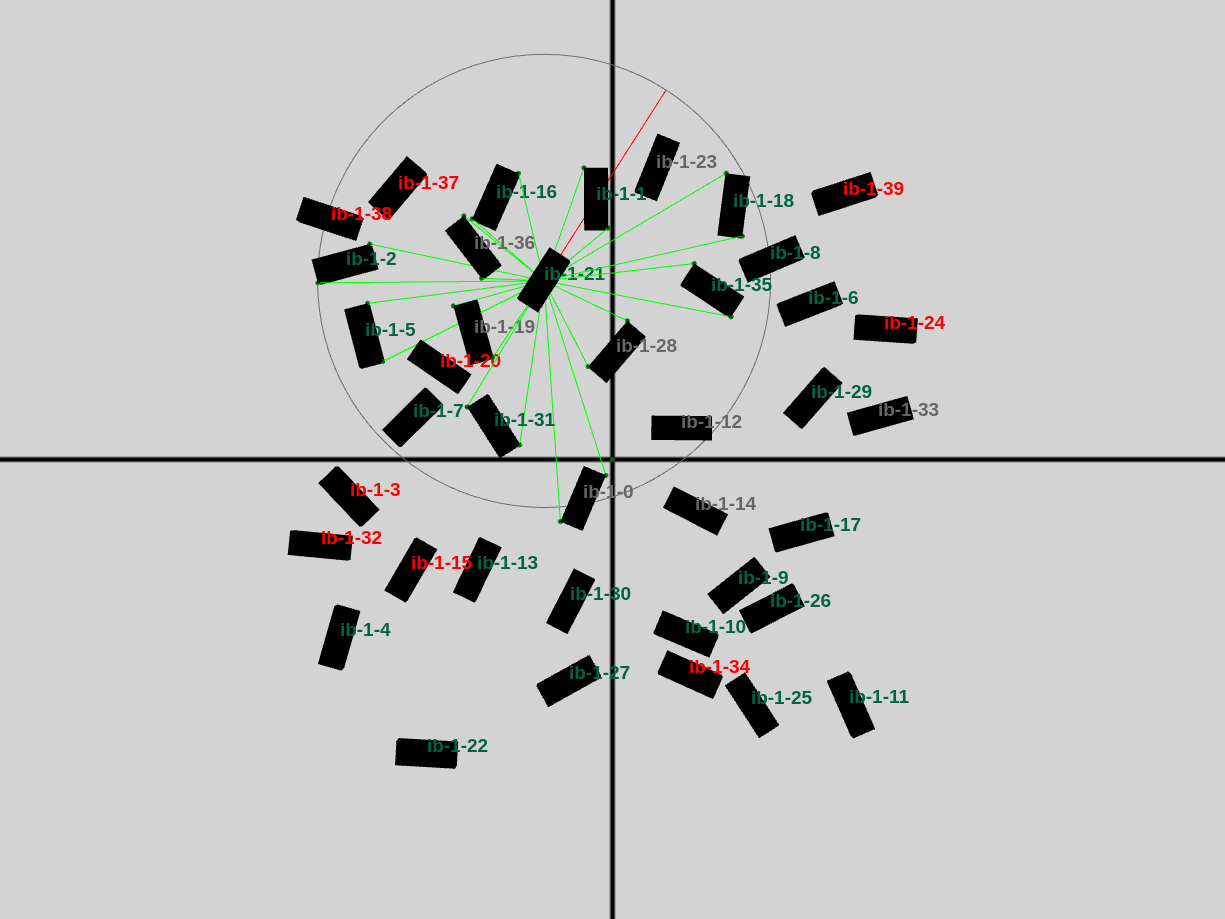}
	\caption{Screenshot from the ARGoS3~\citep{pinciroli12} showing a swarm of 40 simulated \textit{Nymbots}. The gray circle
		represents the sensing radius ($r_{sense}$) of a robot.
}
	\label{fig:arena}
\end{figure}

The experimental arena is effectively infinite, ensuring robots do not reach the boundary during the experiment. The robots are initially positioned at random locations (uniform distribution) in a $0.5\m \times 0.5\m$ area at the center of the arena. Initial headings are randomized as well. \Cref{fig:arena} shows a top-down view of the arena, with 40 robots.

The primary measure of performance we utilize in this paper is the \emph{order parameter} of the collective motion. First introduced by \cite{vicsek95}, the order quantifies the heading alignment of the swarm.  Given the set $\mathcal{A}$ of robots their order is defined as follows:
\begin{equation}
	\text{order}(\mathcal{A}) = \frac{1}{ \left|\mathcal{A} \right|} \sum_{i\in \mathcal{A}} \hat{e_i}  ,
 \label{eq:order}
\end{equation}
where $\hat{e_i}$ denotes the heading of robot $i$ in the global frame of reference.
As the robots become more aligned, the order increases, reaching a value of 1 when all robots are perfectly aligned.
Conversely, if the robots are disordered, the order approaches 0.
Later, when we measure order in swarms included faulty robots that are stationary or too slow, they are not considered in the order parameter, as the goal is to show order is maintained in nominal (healthy) robots.

\begin{figure}
	\centering
	\includegraphics[width=0.99\columnwidth]{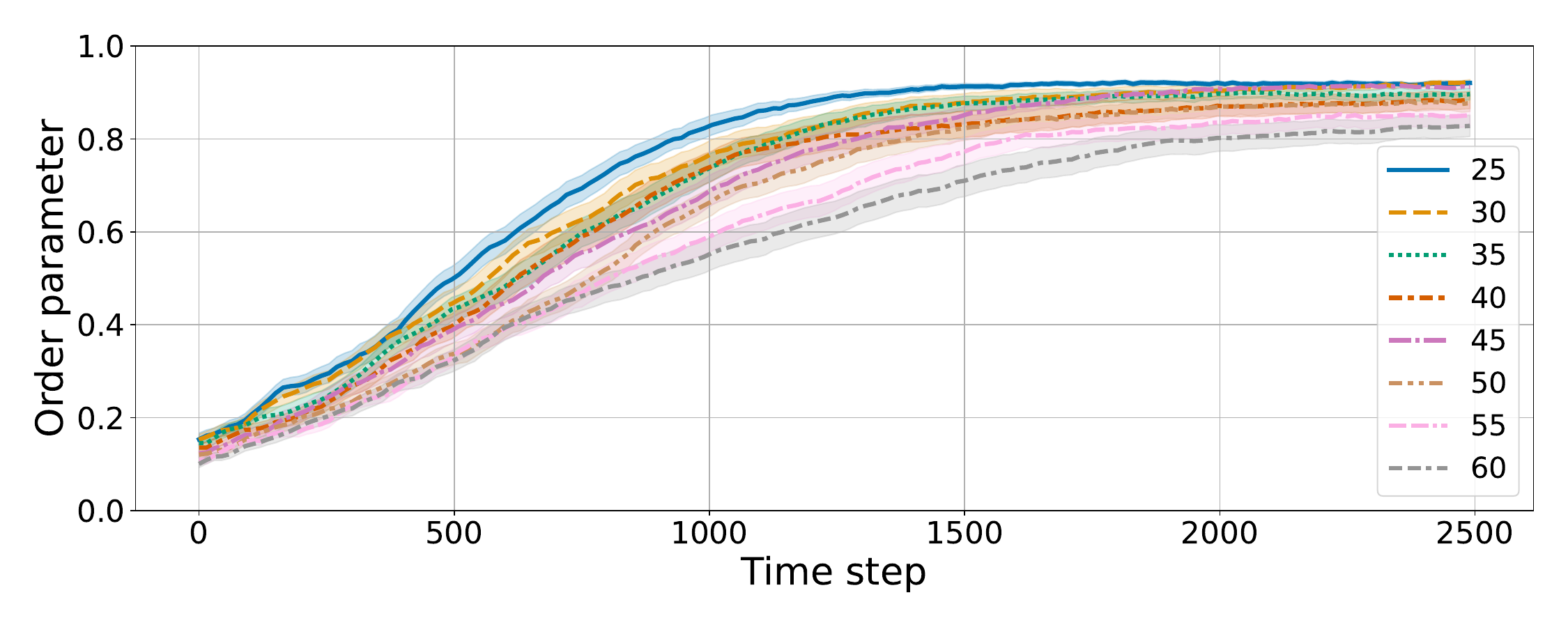}
\caption{Evolving order parameter of the system with different swarm sizes.}
	\label{fig:baseline_order_parameter}

\end{figure}

\Cref{fig:baseline_order_parameter} shows the evolution of order for different swarm sizes.
The parameter values used are detailed in \cref{tab:aa_params_appendix} (\cref{app:aa}).
We will be presenting figures such as this throughout the rest of the paper, and so we explain it in detail here.
The vertical axis measures the order parameter described above.  The horizontal axis measures time-steps (simulation \emph{ticks}), where $1$ tick is equal to 0.1\unit{\second} (i.e., 10 ticks per second). Unless stated otherwise, experiments run for 12,000 ticks (1,200 seconds), and shaded regions in plots denote the standard error over 50 trials.
The figure shows that all swarm sizes (different lines) begin in a disordered state (low order) and converge to an ordered state over time. All swarm sizes eventually reach an ordered state, though larger swarms require more time to converge and achieve a lower order.

Although \emph{order} is the primary measure, we also report on the \emph{speed} of the swarm (when utilizing various models) in later sections. This is to investigate possible side effects of robust motion.

\section{Robust Vision-Based Sensing and Neighbor Detection}
\label{ch:pereception_challenges}

Vision is a fundamental sensory modality in natural agents, yet it is often idealized in theoretical models of collective motion.
In this section, we instantiate the AA model under vision-based sensing for elongated robots and address two core challenges that are raises in the \emph{neighbor detection} stage:
robust estimation of distance to detected robots (discussed in \cref{sec:distance_estimation}) and handling occlusions (\cref{sec:occlusions}). We show how these combine to create a new model (AA-V) that improves on the original (\cref{sec:aav-eval}).

We begin by defining the sensing modality that raises these challenges.
We assume a visual sensor that does not provide depth information.
This occurs naturally with monocular cameras, but is also true in robots (and animals) that combine non-overlapping fields of view into a wide field of view.
In principle, an idealized visual sensor would have infinite resolution (and thus no sensing range limits).
In practice, however, image resolution is finite: an object so distant that its projected size falls below one pixel becomes undetectable.
Thus, an effective sensing range $r_{sense}$ can be established for the sensor.
Even within that range, there are pixelization and quantization effects that would cause
measurement errors. We do not simulate such errors in our experiments for reasons of simplicity.

Every pixel in the image projected on the camera is associated with a vertical and a horizontal imaging angles relative to the focal robot's local coordinate system.
That is, the sensor inherently provides angular information (bearing and elevation) to any visible object, but it does not measure range, nor does it penetrate objects: robots or parts of robots that are occluded by others are not sensed.
We assume the sensor can distinguish whether a pixel is imaging a neighboring robot.

\subsection{Detecting Neighbors in the Presence of Occlusions}
\label{sec:occlusions}
Swarms operate in crowded environments, where robots are frequently subject to occlusions caused by their neighbors.
These occlusions, which may be partial or complete, reduce the information available to a robot as fully occluded robots are not sensed.
Critically, occlusions also
lead to actual misinterpretation of sensed information. For instance, a partially occluded neighbor may be mistaken for a fully visible, more distant robot, leading to incorrect motion decisions.

Occlusions cannot be controlled or eliminated, as they are inherent to the visual sensor.
However, their existence can be addressed in various ways, depending on the assumptions made and the computational complexity involved. For example, if focal robots can visually recognize others (e.g., by a visual object recognition method) then they can distinguish between partially occluded and fully visible neighbors.

We make the assumption that robots can indeed recognize neighbors when they are visible. This is a common assumption in existing research, though several recent  investigations attempt to handle occlusions without this assumed capability~\citep{krongauz24,mezey25,li24}. The assumption defines a spectrum of occlusion handling methods.

At one end of the spectrum is the \emph{OMID} method. Fully visible robots are recognized, and anything else is \emph{omitted}, in particular: partially occluded neighbors are completely ignored and are not detected---and therefore do not influence the focal robot's decision making.

At the other end of the spectrum is the \emph{COMPLID} method, which assumes not only the ability to recognize fully visible robots, but also to cognitively \emph{complete} any partial image of a robot to its full image. Visual information about a partially occluded neighbor is extrapolated to create a complete image of the neighbor, including any hidden parts.

In between these two ends of the spectrum lie a family of variants that can carry out some extrapolation, but only when that the partial image is of sufficient size (e.g., more than half the body is visible) or sufficient detail (can recognize if the front of the robot is visible).
To represent these in the experiments, we employed a \emph{CENTER} method, which allows robots to recognize each other if their center is visible.

\begin{figure}[htbp]
	\centering
	\includegraphics[width=0.99\linewidth]{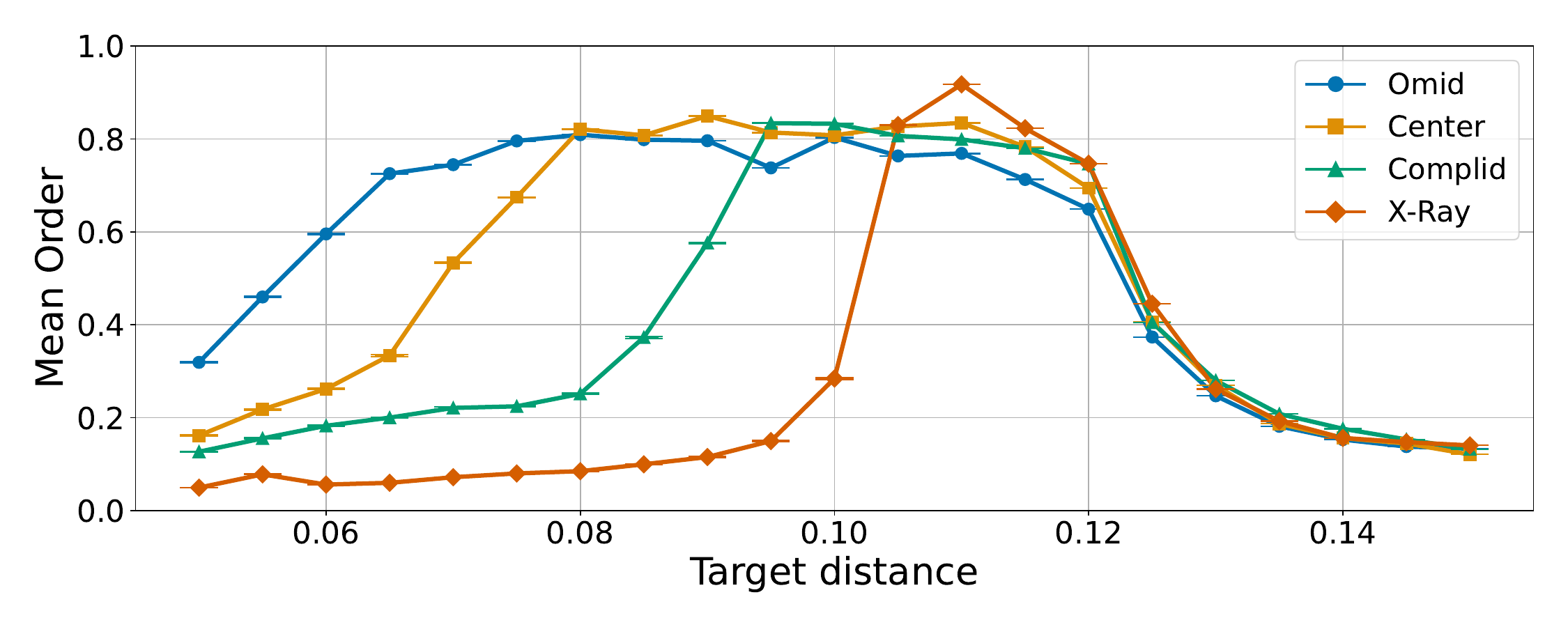}
	\caption{Performance comparison of occlusion handling methods across varying $r_d$ and fixed $r_{sense}=0.19\m$. Experiments were carried out in a swarm of 40 robots, using the AA model. Means displayed are across 50 trials per setting. Occlusion-aware models exhibit reduced sensitivity to $r_d$ variations, maintaining more stable collective behavior.}

	\label{fig:sensitive_distance}
\end{figure}

\Cref{fig:sensitive_distance} compares the order parameter achieved (at steady state) by use of the AA method, using the different occlusion-handling methods. The order parameter achieved is plotted across different $r_d$ values while keeping $r_{sense}$ constant at 0.19\m. The figure contrasts the occlusion handling methods presented above, adding also the baseline idealized method (\emph{X-RAY}), which always detects all robots within the sensing range. This is the method often assumed by theoretical models of collective vision. Naturally, it is impossible to use with a visual sensor in practice.

\Cref{fig:sensitive_distance} reveals that occlusion methods shows differ in their sensitivity to the $r_d$ selected. The X-RAY method (red, diamond markers) achieves the highest order but is highly sensitive, achieving high order only for $r_d\in [0.105,0.12]$ (approximately).

The COMPLID method (green, triangle markers) is the closest visually-feasible approximation to the X-RAY method.
It is less sensitive to $r_d$ values. It maintains high order for $r_d\in [0.095, 0.12]$.

As we move on the spectrum from the COMPLID end to the OMID end, sensitivity to $r_d$ decreases. The CENTER method (orange, square markers) works well with $r_d\in [0.08,0.12]$, and the OMID method (blue, circle markers) works well for $r_d\in [0.07,0.115]$.  All occlusion-aware methods reach peak performance at lower $r_d$ values than \textit{X-RAY}. As occlusions exclude distant neighbors, weakening attraction forces and increasing the risk of robots disconnecting. Lowering $r_d$ mitigates this by encouraging tighter formations.

The peak performance of the methods is essentially identical when \(r_d\) is chosen appropriately, suggesting the choice of occlusion handling may be inconsequential. Yet, distance estimation relies on assumed neighbor dimensions; variation in size introduces proportional errors (e.g., a robot \(5\%\) taller appears \(5\%\) closer). It may therefore be preferable to select methods more robust to variance in \(r_d\). As our experiments use homogeneously sized robots, we leave this question to future work and, unless otherwise noted, adopt \textsc{COMPLID} for the experiments reported below.

\subsection{Distance and Bearing}
\label{sec:distance_estimation}
Regardless of how occlusions are handled, the output from the perceptual process that recognizes members is a set of neighbors $\mathcal{D}_i^t$. For each such member $j\in\mathcal{D}_i^t$, the avoid-attract force vector (\cref{eq:rjvector}) requires knowledge of a vector $\Ct{}=(r_{ij},\angle \beta_{ij})$, pointing
to the center of the neighbour $j$, in polar coordinates centered in the focal robot $i$ and with respect to its heading.

However, because a realistic visual sensor does not directly sense range, nor identify the geometric center of a neighbor, $\Ct{}$ must be estimated from
the visual information. To do this, we will be using the two outermost vertical edges of the each box-like neighbor (see below), and the known physical dimensions of the rectangular (box) robots. These are denoted $l$ (length), $w$ (width), and  $v$ (vertical size/height).

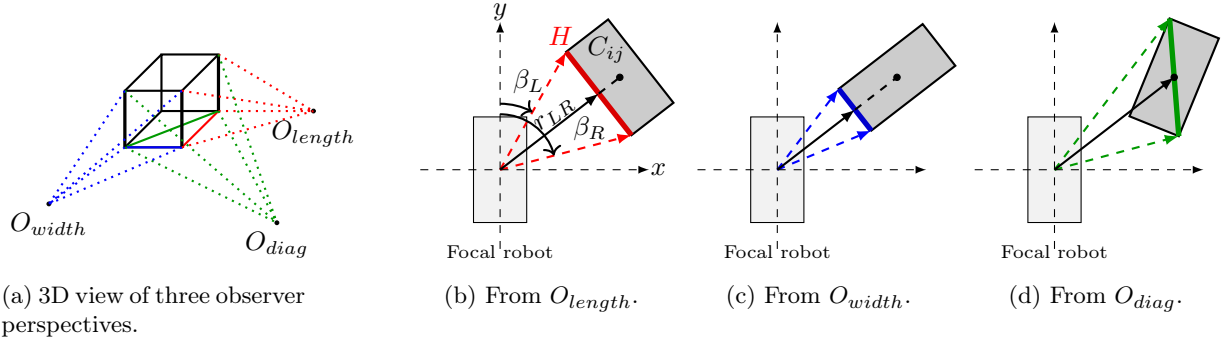
\begin{figure*}[htbp]
\begin{subfigure}[t]{0.25\textwidth}
		\centering
		\begin{tikzpicture}[scale=0.25]
\coordinate (A) at (5,0,0); \coordinate (B) at (8,0,0);
			\coordinate (C) at (8,3,0); \coordinate (D) at (5,3,0);
			\coordinate (E) at (5,0,5); \coordinate (F) at (8,0,5);
			\coordinate (G) at (8,3,5); \coordinate (H) at (5,3,5);

\coordinate (V)  at (13,0,0);
			\coordinate (V1) at (13,-4,5);
			\coordinate (V2) at (1,-3,5);
			\filldraw[black] (V)  circle (0.1) node[below] {$O_{length}$};
			\filldraw[black] (V1) circle (0.1) node[below] {$O_{diag}$};
			\filldraw[black] (V2) circle (0.1) node[below] {$O_{width}$};

\begin{scope}[shift={(8,0)}]
\draw[thick] (A) -- (B) -- (C) -- (D) -- cycle;
				\draw[thick] (E) -- (F) -- (G) -- (H) -- cycle;
				\draw[thick] (A) -- (E);
				\draw[thick,red]   (B) -- (F);
				\draw[thick,green!60!black] (E) -- (B);
				\draw[thick,blue](E) -- (F);
				\draw[thick] (C) -- (G); \draw[thick] (D) -- (H);
				\draw[very thick,black](E) -- (H); \draw[very thick,black](F) -- (G);
				\draw[very thick,black](C) -- (B);

\draw[red, thick, dotted] (B)--(V) (C)--(V) (F)--(V) (G)--(V);
				\draw[green!60!black, thick, dotted] (B)--(V1) (C)--(V1) (E)--(V1) (H)--(V1);
				\draw[blue, thick, dotted] (G)--(V2) (F)--(V2) (E)--(V2) (H)--(V2);
			\end{scope}
		\end{tikzpicture}
		\caption{3D view of three observer perspectives.}
		\label{fig:robot_axis_system}
	\end{subfigure}
\pgfmathsetmacro{\Mx}{2.25}
	\pgfmathsetmacro{\My}{1.75}
	\pgfmathsetmacro{\aW}{0.5}  \pgfmathsetmacro{\bL}{1.0}  \pgfmathsetmacro{\angM}{atan2(\My,\Mx)}
	\pgfmathsetmacro{\ux}{cos(\angM)}
	\pgfmathsetmacro{\uy}{sin(\angM)}
	\pgfmathsetmacro{\vx}{-sin(\angM)}
	\pgfmathsetmacro{\vy}{cos(\angM)}
\hfill
	\begin{subfigure}[t]{0.2\textwidth}
		\centering
		\begin{tikzpicture}[scale=0.7]
\filldraw[fill=gray!10] (-0.5,1) rectangle (0.5,-1);

\coordinate (O) at (0,0);
			\coordinate (M) at (\Mx,\My);

\pgfmathsetmacro{\aUse}{\bL}
			\pgfmathsetmacro{\bUse}{\aW}

			\coordinate (P1) at ({\Mx - \bUse*\ux + \aUse*\vx},{\My - \bUse*\uy + \aUse*\vy});
			\coordinate (P2) at ({\Mx - \bUse*\ux - \aUse*\vx},{\My - \bUse*\uy - \aUse*\vy});
			\coordinate (Q1) at ({\Mx + \bUse*\ux + \aUse*\vx},{\My + \bUse*\uy + \aUse*\vy});
			\coordinate (Q2) at ({\Mx + \bUse*\ux - \aUse*\vx},{\My + \bUse*\uy - \aUse*\vy});

\draw[thick, fill=gray!40] (P1)--(P2)--(Q2)--(Q1)--cycle;

\draw[-latex, thick, red, dashed] (O) -- (P1);
			\draw[-latex, thick, red, dashed] (O) -- (P2);

\draw[red!85!black, line width=2pt] (P1)--(P2);

\node[inner sep=1pt, red] at ($(P1)+(-0.1,0.3)$) {$H$};

\draw[-latex, dashed] (0,-1.5) -- (0,2.8) node[above, fill=white, inner sep=0.5pt] {$y$};
			\draw[-latex, dashed] (-1.5,0) -- (2.8,0) node[right, fill=white, inner sep=0.5pt] {$x$};
			\node at (0,-1.55) {\footnotesize Focal robot};

\coordinate (Cedge) at ($ (P1)!0.5!(P2) $);   \coordinate (Cmid)  at ($ (P1)!0.5!(Q2) $);

\filldraw[draw=black, fill=black] (Cmid) circle (0.06);
			\node[above=2pt, xshift=-5pt] at (Cmid) {$C_{ij}$};

\draw[-latex, thick, black] (O) -- (Cedge)
			node[pos=0.62, sloped, above=0pt, inner sep=0.5pt] {$r_{LR}$};

\draw[thick, black, dashed] (Cedge) -- (Cmid);

\pgfmathsetmacro{\angStart}{90}

\pgfmathanglebetweenpoints{\pgfpoint{0cm}{0cm}}{\pgfpointanchor{P1}{center}}
			\let\angL\pgfmathresult
			\pgfmathanglebetweenpoints{\pgfpoint{0cm}{0cm}}{\pgfpointanchor{P2}{center}}
			\let\angR\pgfmathresult

\pgfmathsetmacro{\deltaCWL}{mod(\angStart - \angL,360)}
			\pgfmathsetmacro{\deltaCWR}{mod(\angStart - \angR,360)}
			\pgfmathsetmacro{\angEndL}{\angStart - \deltaCWL}
			\pgfmathsetmacro{\angEndR}{\angStart - \deltaCWR}

\pgfmathsetmacro{\RbetaL}{1.25}
			\pgfmathsetmacro{\RbetaR}{1.05}
			\pgfmathsetmacro{\angLabL}{\angStart - 0.55*\deltaCWL}
			\pgfmathsetmacro{\angLabR}{\angStart - 0.55*\deltaCWR}

\path (O) ++(\angStart:\RbetaL) coordinate (betaStartL);
			\draw[thick,->] (betaStartL) arc[start angle=\angStart, end angle=\angEndL, radius=\RbetaL];
			\node[font=\bfseries] at ($(O)+({(\RbetaL+0.7)*cos(\angLabL)},{(\RbetaL+0.50)*sin(\angLabL)})$) {$\beta_L$};

\path (O) ++(\angStart:\RbetaR) coordinate (betaStartR);
			\draw[thick,->] (betaStartR) arc[start angle=\angStart, end angle=\angEndR, radius=\RbetaR];
			\node[font=\bfseries] at ($(O)+({(\RbetaR+1.5)*cos(\angLabR)},{(\RbetaR+0.00)*sin(\angLabR)})$) {$\beta_R$};

		\end{tikzpicture}
		\caption{From $O_{length}$.}
		\label{fig:topdown-length}
	\end{subfigure}
\hfill
	\begin{subfigure}[t]{0.2\textwidth}
		\centering
		\begin{tikzpicture}[scale=0.7]

\filldraw[fill=gray!10] (-0.5,1) rectangle (0.5,-1);

\coordinate (O) at (0,0);
			\coordinate (M) at (\Mx,\My);

\coordinate (P1) at ({\Mx - \bL*\ux + \aW*\vx},{\My - \bL*\uy + \aW*\vy});
			\coordinate (P2) at ({\Mx - \bL*\ux - \aW*\vx},{\My - \bL*\uy - \aW*\vy});
			\coordinate (Q1) at ({\Mx + \bL*\ux + \aW*\vx},{\My + \bL*\uy + \aW*\vy});
			\coordinate (Q2) at ({\Mx + \bL*\ux - \aW*\vx},{\My + \bL*\uy - \aW*\vy});

\draw[thick, fill=gray!40] (P1)--(P2)--(Q2)--(Q1)--cycle;

\draw[-latex, thick, blue, dashed] (O) -- (P1);
			\draw[-latex, thick, blue, dashed] (O) -- (P2);

\draw[blue!80!black, line width=2pt] (P1)--(P2);

\draw[-latex, dashed] (0,-1.5) -- (0,2.8);
			\draw[-latex, dashed] (-1.5,0) -- (2.8,0);
			\node at (0,-1.55) {\footnotesize Focal robot};

\coordinate (Cedge) at ($ (P1)!0.5!(P2) $);   \coordinate (Cmid)  at ($ (P1)!0.5!(Q2) $);

\filldraw[draw=black, fill=black] (Cmid) circle (0.06);

\draw[-latex, thick, black] (O) -- (Cedge);

\draw[thick, black, dashed] (Cedge) -- (Cmid);

\pgfmathsetmacro{\angStart}{90} \pgfmathanglebetweenpoints{\pgfpoint{0cm}{0cm}}{\pgfpointanchor{Cmid}{center}}
			\let\angC\pgfmathresult
			\pgfmathsetmacro{\deltaCW}{mod(\angStart - \angC,360)}
			\pgfmathsetmacro{\angEnd}{\angStart - \deltaCW}
			\pgfmathsetmacro{\Rbeta}{1.25} \pgfmathsetmacro{\angLab}{\angStart - 0.55*\deltaCW}

		\end{tikzpicture}
		\caption{From $O_{width}$.}
		\label{fig:topdown-width}
	\end{subfigure}
	\hfill
\begin{subfigure}[t]{0.2\textwidth}
		\centering
		\begin{tikzpicture}[scale=0.7]
\filldraw[fill=gray!10] (-0.5,1) rectangle (0.5,-1);

\pgfmathsetmacro{\Mx}{2.25}
			\pgfmathsetmacro{\My}{1.75}
			\pgfmathsetmacro{\aW}{0.5}
			\pgfmathsetmacro{\bL}{1.0}
			\coordinate (O) at (0,0);
			\coordinate (M) at (\Mx,\My);
			\pgfmathsetmacro{\angM}{atan2(\My,\Mx)}
			\pgfmathsetmacro{\ux}{cos(\angM)}
			\pgfmathsetmacro{\uy}{sin(\angM)}
			\pgfmathsetmacro{\vx}{-sin(\angM)}
			\pgfmathsetmacro{\vy}{cos(\angM)}

\begin{scope}[rotate around={30:(\Mx,\My)}]
				\coordinate (P1) at ({\Mx - \bL*\ux + \aW*\vx},{\My - \bL*\uy + \aW*\vy}); \coordinate (P2) at ({\Mx - \bL*\ux - \aW*\vx},{\My - \bL*\uy - \aW*\vy}); \coordinate (Q1) at ({\Mx + \bL*\ux + \aW*\vx},{\My + \bL*\uy + \aW*\vy}); \coordinate (Q2) at ({\Mx + \bL*\ux - \aW*\vx},{\My + \bL*\uy - \aW*\vy});

\draw[thick, fill=gray!40] (P1)--(P2)--(Q2)--(Q1)--cycle;
				\coordinate (Cmid)  at ($ (P1)!0.5!(Q2) $);

\draw[-latex,dashed, thick, green!60!black] (O) -- (P2);
				\draw[-latex,dashed, thick, green!60!black] (O) -- (Q1);
				\draw[green!60!black, line width=2.0pt] (P2)--(Q1);
				\filldraw[draw=black, fill=black] (Cmid) circle (0.06);

\end{scope}

\draw[-latex, dashed] (0,-1.5) -- (0,2.8);
			\draw[-latex, dashed] (-1.5,0) -- (2.8,0);
			\node at (0,-1.55) {\footnotesize Focal robot};

\coordinate (MidEdge) at ($ (Q1)!0.5!(P2) $);
			\draw[-latex, thick, black] (O) -- (MidEdge);
		\end{tikzpicture}
		\caption{From $O_{diag}$.}
		\label{fig:topdown-diag}
	\end{subfigure}

	\caption{Neighbor views from different observing positions. \textbf{(a)} 3D rays to the outermost horizontal and vertical edges. \textbf{(b)--(d)} 2D top-down views of the horizontal effective edges, with notation as in \cref{fig:topdown-length}.}

	\label{fig:topdown}
\end{figure*}

\Cref{fig:robot_axis_system} shows how the visual sensor perceives a neighbor from different focal robot observation positions (denoted $O$).
The two outermost vertical edges  from each perspective (bold lines) are denoted $L$ and $R$.
In the top-down views (\cref{fig:topdown-length,fig:topdown-width,fig:topdown-diag}) the horizontal positions of the edges are pointed by the vectors $\Lt{},\Rt{}$ (defined next).

Let $\beta_L$ and $\beta_R$ be the bearing angles to the vertical edges $L,R$ (resp.).
The edges define the widest horizontal subtended angle $\theta = \beta_R - \beta_L$, which is also known as the angular width of the neighbor.

Two vertical angles $\alpha_L,\alpha_R$ are subtended on the two vertical edges $L,R$ (resp.).  Using the known vertical size of the neighbor ($v$), we compute the distances $r_R$ and $r_L$ for the right and left edges respectively, using $\alpha\in \{\alpha_R,\alpha_L\}$:
\begin{equation}
	r_\alpha \;=\; \frac{1}{2}\,v \,\cot\!\left(\frac{\alpha}{2}\right)
	\label{eq:r_vertical}
\end{equation}

We can now define the vectors
\[
\vec{L} = (r_L,\ \beta_L), \qquad
\vec{R} = (r_R,\ \beta_R).
\]
They are illustrated in \cref{fig:topdown-length,fig:topdown-width,fig:topdown-diag}.
By construction, these vectors lie in a plane parallel to the ground, and  orthogonal to the vertical edges.
We use the two vectors $\Lt{},\Rt{}$ (separately for each neighbor) to
an estimated $\Ct{}$ (\cref{eq:cvector}), constructed from visual perception.

\paragraph{Estimating the Relative Bearing Angle}
The bearing angle $\beta_C$ to the center of the neighbor is
\begin{equation}
\beta_C = \frac{\beta_R+\beta_L}{2}.
\end{equation}
This requires only angular information, independent of distance, and thus provides a robust estimator under monocular vision.

\paragraph{Estimating the Distance}
The estimation of range $r_C$ to the center of the neighbor is more involved. While subtended angles $\theta$, $\alpha_L$ and $\alpha_R$ can be used to estimate range, they differ in reliability.

\begin{figure}[htbp]
	\centering
	\begin{tikzpicture}[scale=0.9]
\filldraw[black] (0,2) circle (0.08);
		\node[below] at (0,1.9) {\footnotesize Observer};

\draw[fill=gray!40] (5.0,3.0) -- (5.0,1.0) -- (5.75,1.0) -- (5.75,3.0) -- cycle;

\begin{scope}[rotate around={90:(3.15,1.6)}]
			\draw[fill=gray!40, opacity=0.7] (3.15,3.0) -- (3.15,1.0) -- (3.9,1.0) -- (3.9,3.0) -- cycle;
		\end{scope}

\draw[dashed] (0,2) -- (5.0,1.0);
		\draw[dashed] (0,2) -- (5.0,3.0);
	\end{tikzpicture}
	\caption{Illustration of distance estimation errors that may occur when using the angular width. This is a top-down view of two robots at different distances and orientations subtending the same angle at the observer (same angular width).}
	\label{fig:subtended_angle_vs_distances}
\end{figure}
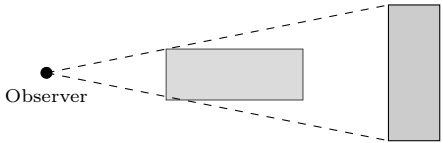

The horizontal subtended angle $\theta$ is a poor estimator~\citep{bastien20,mezey25,krongauz24}. Two neighbors at very different distances may project the same angular width if their orientations differ (\cref{fig:subtended_angle_vs_distances}).

The vertical subtended angles are generally more robust, as they are largely orientation invariant~\citep{moshtag09,ito24,harpaz21}.  Because of this,
the range at the center of the robot can be estimated as
\begin{equation}
	r_{LR} \;=\; \frac{r_L+r_R}{2}\;.
\label{eq:r_lr}
\end{equation}

However, even this estimator introduces errors: it yields the distance to the \emph{facing edge} of the neighbor rather than its center.
This can be seen in~\cref{fig:topdown-length,fig:topdown-width}: the solid line
marking $r_{LR}$ reaches the center of the wide side or long side of the neighbor, but not its center.  In both cases, a correction is needed (dashed line continuing the center line segment).

To make this correction, we need to calculate the corrective offset from the edge to the center (denoted $\upsilon_C$). To do this, the focal robot requires the metric length of the visible horizontal edge. This can be used to determine whether the robot is facing the width of the neighbor (\cref{fig:topdown-width}), or perhaps its length (\cref{fig:topdown-length}).

The metric length of the visible horizontal edge is the norm of their difference evaluated in rectangular (Cartesian) coordinates. We slightly abuse the standard notation to denote this (all vectors are in polar form):
\begin{equation}
  H = \left\| \Rt{} - \Lt{} \right\|.
\end{equation}

We then compare $H$ against the known physical dimensions of the neighbor ($l,w$). In the general case, the observed span corresponds to the diagonal between two opposing corners, and the distance estimate remains $r_{LR}$. Only in the special cases where the neighbor is oriented broadside does $H$ align with a pure side: if it matches the robot’s width, the center lies half a width deeper than $r_{LR}$ (\cref{fig:topdown-length}); if it matches the robot’s length, the center lies half a length deeper (\cref{fig:topdown-width}). Formally, the correction is:

\begin{equation}
  \upsilon_C =
  \begin{cases}
    \tfrac{w}{2}, & \text{if } \big|H - l\big| \le \varepsilon, \\[6pt]
    \tfrac{l}{2}, & \text{if } \big|H - w\big| \le \varepsilon, \\[6pt]
    0,            & \text{otherwise},
  \end{cases}
\end{equation}
where $\varepsilon$ allows for small estimation errors due to measurement errors,
sensor pixelization and quantization distortions, etc.
The corrected distance is then
\begin{equation}
  r_C = r_{LR} + \mathrm{\upsilon_C}.
  \label{eq:center_correction}
\end{equation}

Hence the final form of the polar vector $\Ct{}$ is:
\begin{equation}
\Ct{} = (r_C,\; \beta_C).
\label{eq:aav}
\end{equation}

\begin{figure}[htbp]
	\centering
	\includegraphics[width=0.99\linewidth]{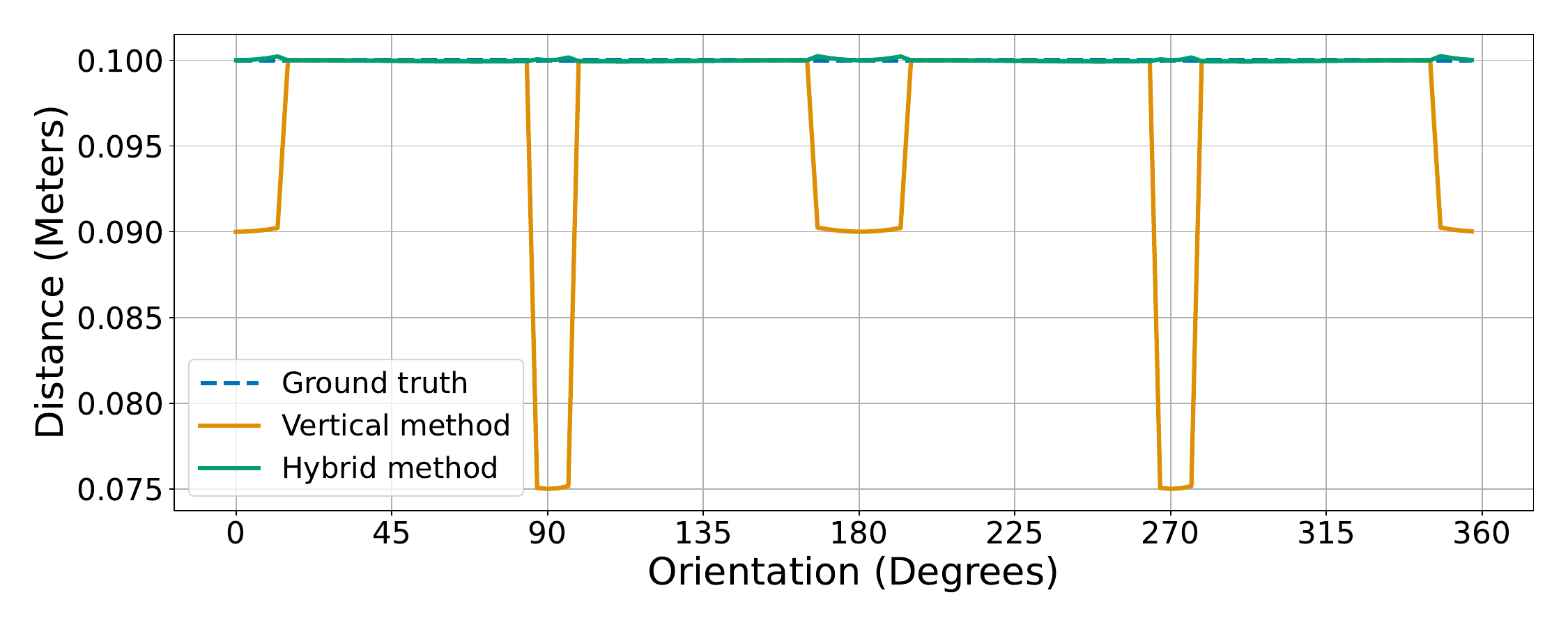}
	\caption{Distance estimates using the vertical angle alone ($r_{LR}$, \emph{Vertical method}) and both vertical and horizontal sizes combined ($r_{C}$, \emph{Hybrid method}). The ground truth is 0.1 m.}
	\label{fig:distance_errors_combined}
\end{figure}

\Cref{fig:distance_errors_combined} plots the distance estimates for an observed Nymbot robot whose true range from the observer is 0.1\m. The estimated distance is plotted as a function of the relative orientation angle.
The figure compares the estimated distance using the vertical effective edge \cref{eq:r_vertical} to
the estimate produced by \cref{eq:center_correction}.  Later (in~\cref{sec:aav-eval}), we show the impact of these estimates on the order parameter.

\subsection{AA-V Collective Motion}
\label{sec:aav-eval}
\begin{figure*}
        \centering
\begin{subfigure}{0.48\textwidth}
            \centering
\includegraphics[width=\columnwidth]{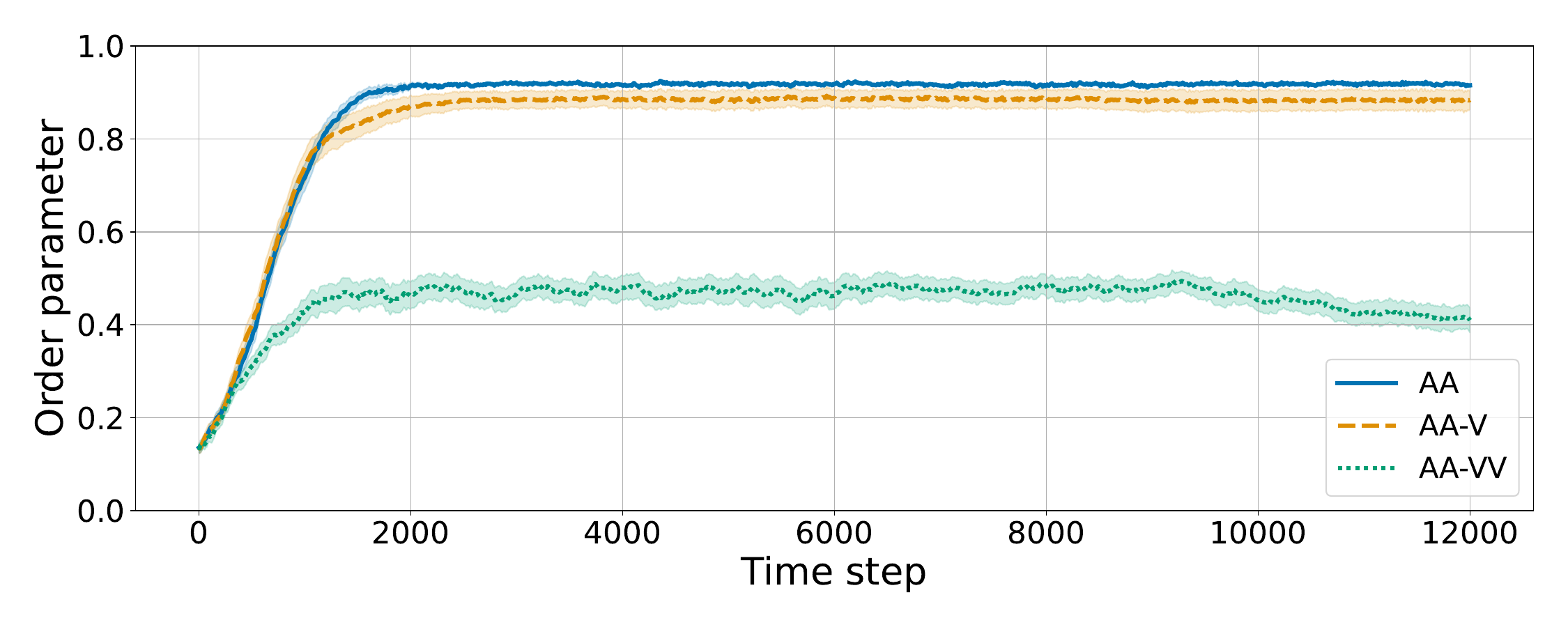}
            \caption{Baseline (X-RAY) $r_d=0.11\m$.}
            \label{fig:vision_x_ray}
        \end{subfigure}
\hfill
\begin{subfigure}{0.48\textwidth}
            \centering
\includegraphics[width=\columnwidth]{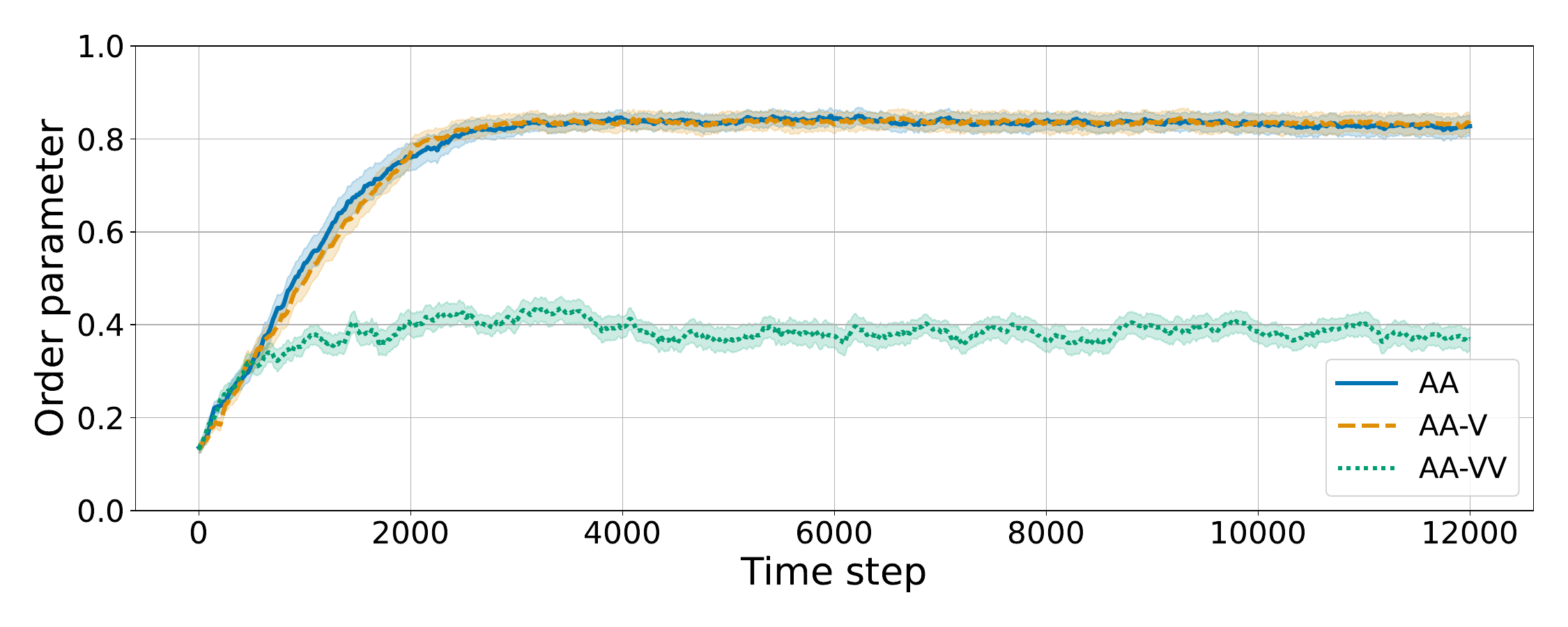}
            \caption{COMPLID. $r_d=0.10\m$.}
            \label{fig:vision_complid}
        \end{subfigure}

        \begin{subfigure}{0.48\textwidth}
			\centering
\includegraphics[width=\columnwidth]{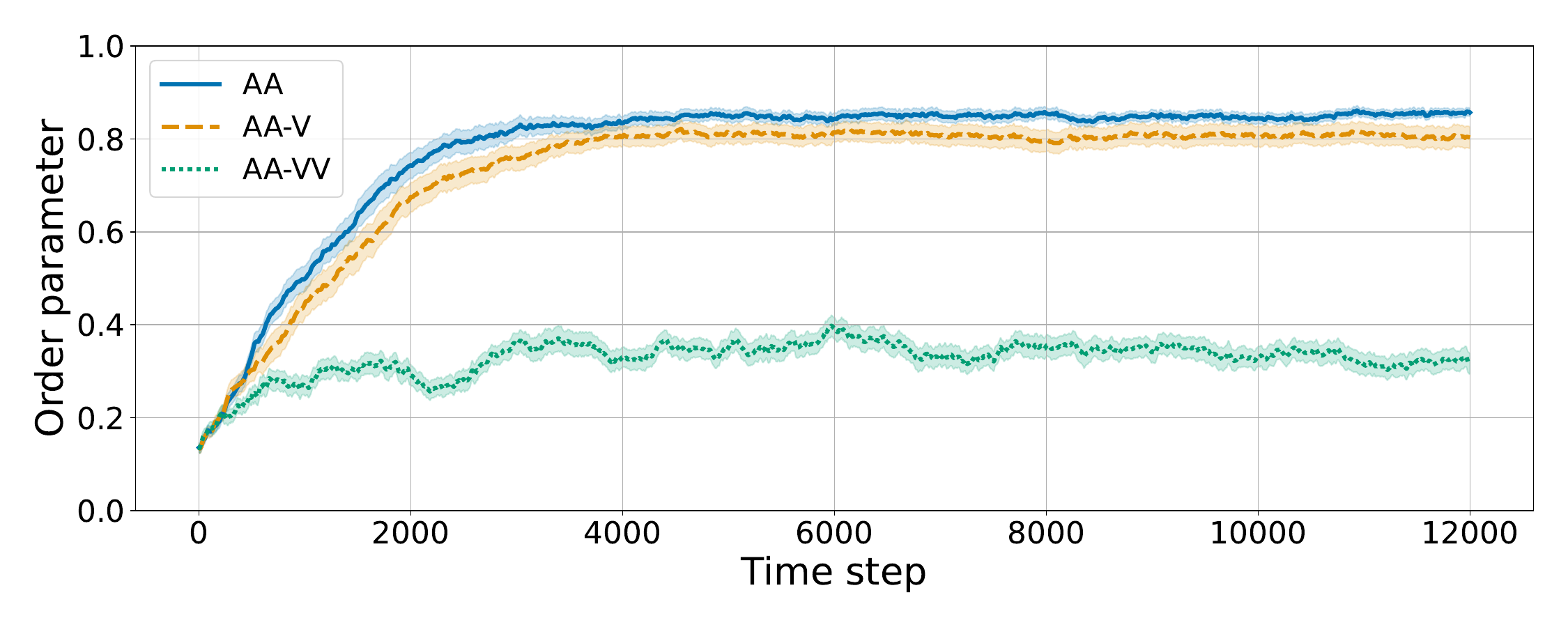}
			\caption{CENTER. $r_d=0.09\m$.}
			\label{fig:vision_center}
		\end{subfigure}
\hfill
        \begin{subfigure}{0.48\textwidth}
            \centering
\includegraphics[width=\columnwidth]{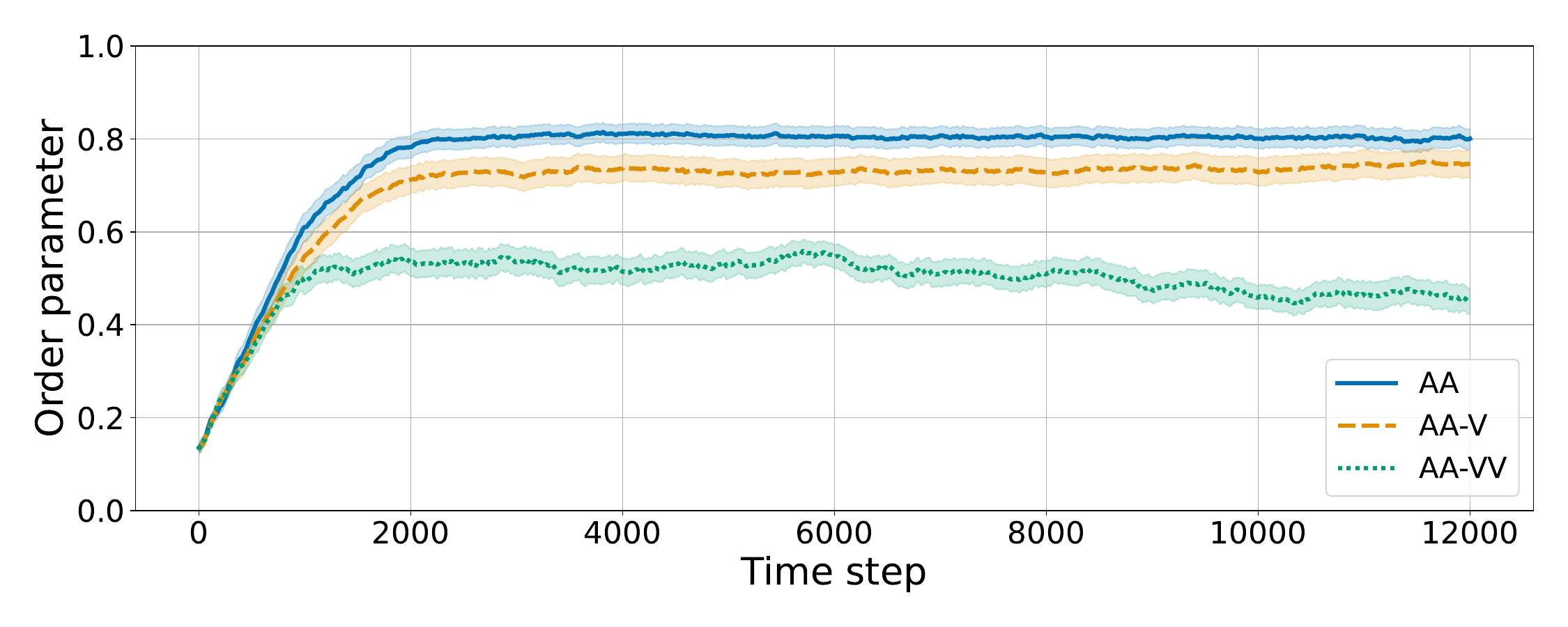}
            \caption{OMID. $r_d=0.10\m$.}
            \label{fig:vision_omid}
        \end{subfigure}

        \caption{Final results of vision for different occlusion handling methods. For each method, the best
        performing $r_d$ value is shown. 40 Robots.}
        \label{fig:vision}

\vspace{48pt}        
\end{figure*}

We now turn to evaluating the impact of visual sensing on neighbor detection, integrating the various methods discussed above.
\Cref{fig:vision} shows the evolution of the order parameters in swarms of 40 robots, with three alternative models:
the baseline AA model uses perfect distance information, and is therefore not possible for use in practice with a visual sensor; the AA-VV model uses $r_{LR}$, \cref{eq:r_lr}; and the AA-V model uses $r_C$, \cref{eq:center_correction}.
All three models are shown in \cref{fig:vision}, operating with the collision handling methods discussed above.

In all cases, the combined vertical/horizontal distance estimation method maintains a high level of collective order, nearly matching the performance of the baseline model. In contrast, relying solely on the vertical method results in lower order, as the estimation errors are both larger and more abrupt.

\section{Selecting Nominal Neighbors}
\label{ch:problem_analysis}
Animals and robots may experience failures that slow or completely stop their locomotion. As we show, and consistent with prior work~\citep{bjerknes13,shefi24}, the presence of such faulty robots leads to a breakdown in the collective motion
generated by both the AA and AA-V (which uses visual sensing). This is because the effects of the relative motion of neighbors with respect to the focal robot are ambiguously captured by vision sensors: the relative motion of a neighbor, as it is captured in a sequence of images over time, is a combination of both the focal robot's velocity as well as the neighbor's. The source velocities leading to the observed relative motion are not distinguished. As a result, faulty neighbors are not isolated from the decision-making for any focal robot detecting them. The faulty robots act as anchors,  nominal robots to cluster around them.

\begin{figure*}[htp]
    \centering
    \begin{subfigure}[b]{0.48\textwidth}
        \centering
        \includegraphics[width=\columnwidth]{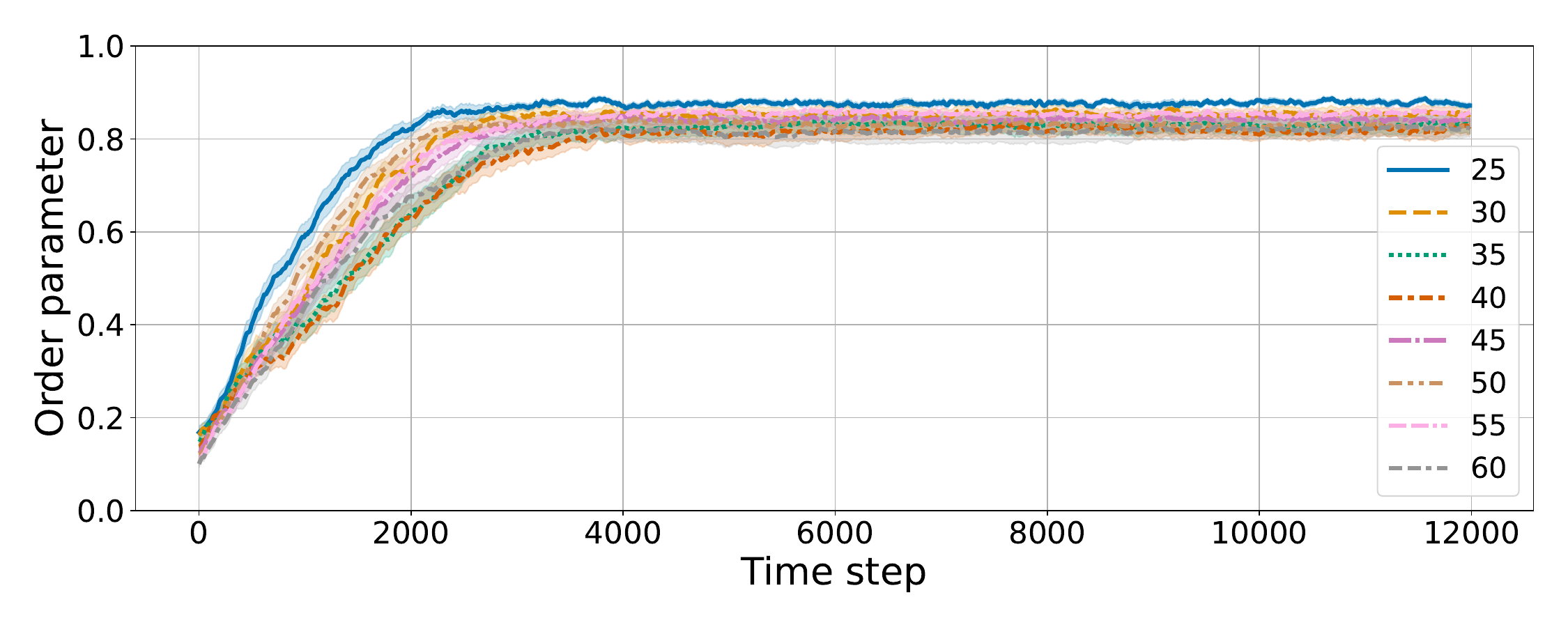}
        \caption{5\% Faulty robots}
        \label{fig:5_faulty}
    \end{subfigure}
    \begin{subfigure}[b]{0.48\textwidth}
        \centering
        \includegraphics[width=\columnwidth]{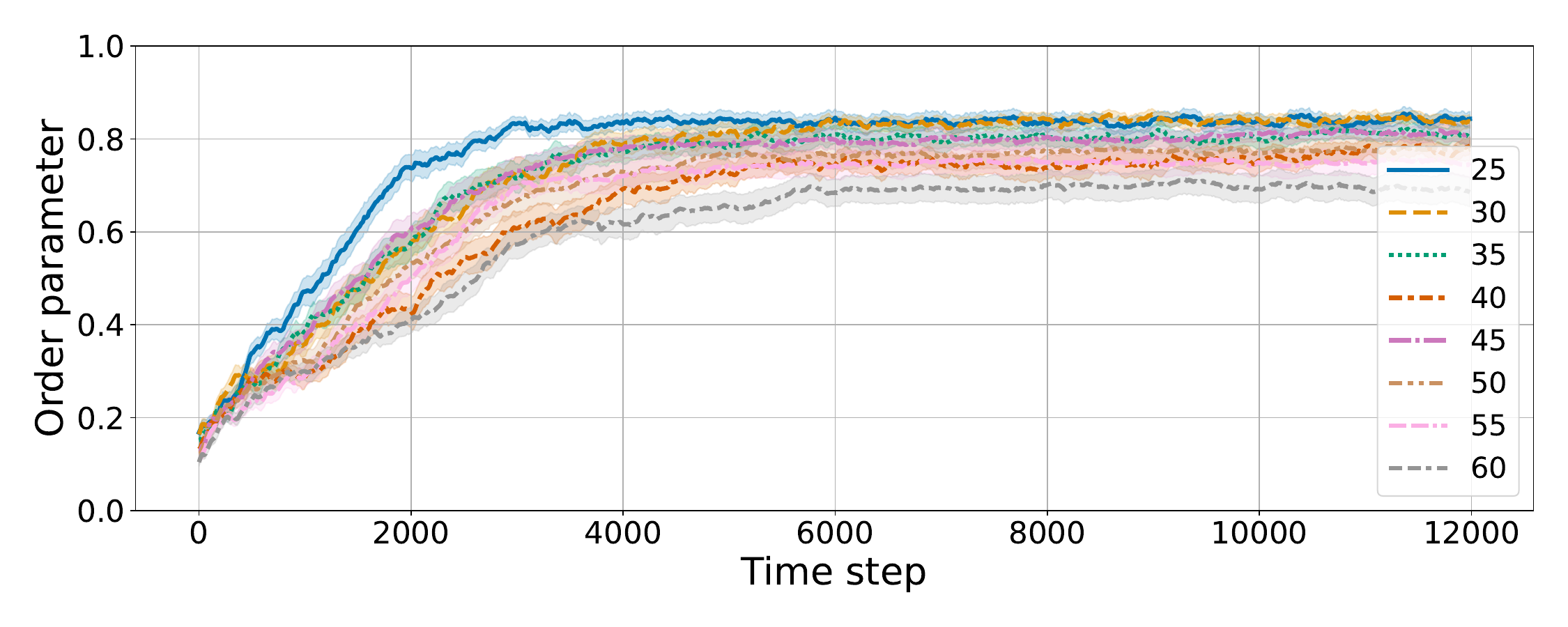}
        \caption{10\% Faulty robots}
        \label{fig:10_faulty}
    \end{subfigure}
    \begin{subfigure}[b]{0.48\textwidth}
        \centering
\includegraphics[width=\columnwidth]{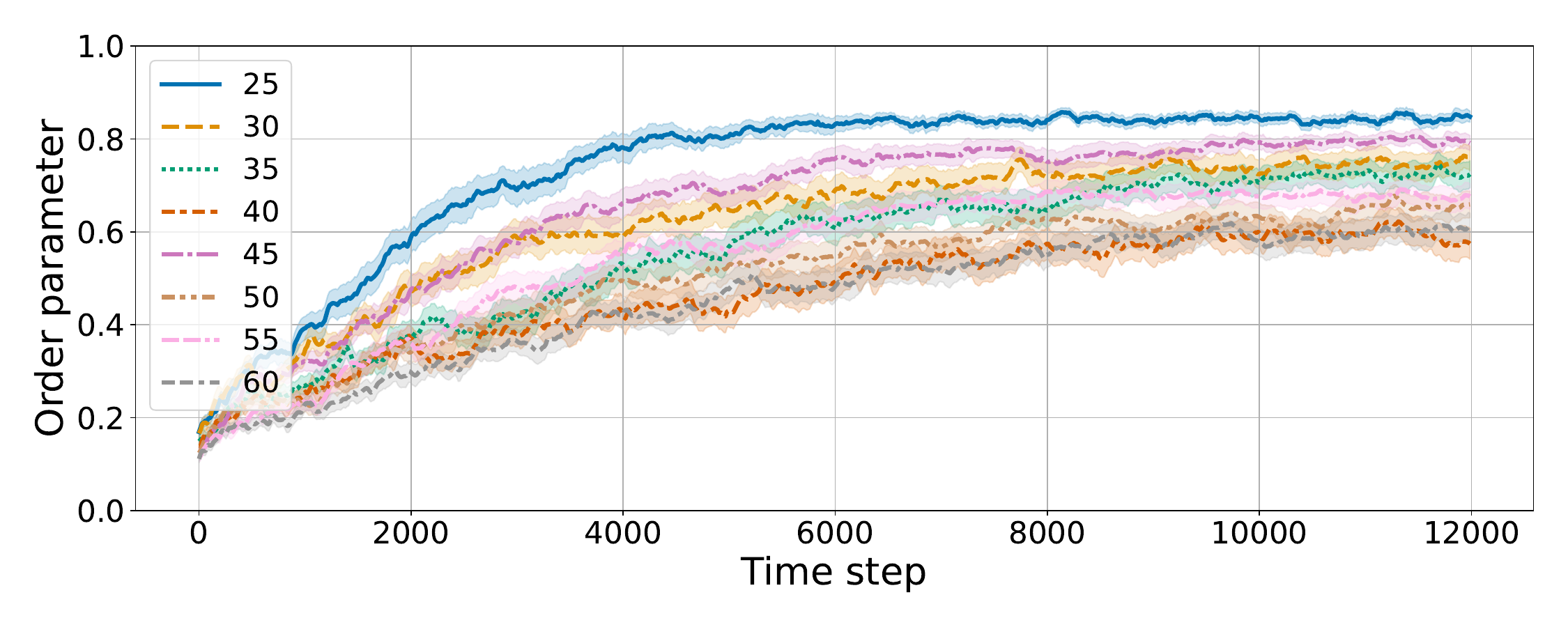}
        \caption{15\% Faulty robots}
        \label{fig:15_faulty}
    \end{subfigure}
    \begin{subfigure}[b]{0.48\textwidth}
        \centering
\includegraphics[width=\columnwidth]{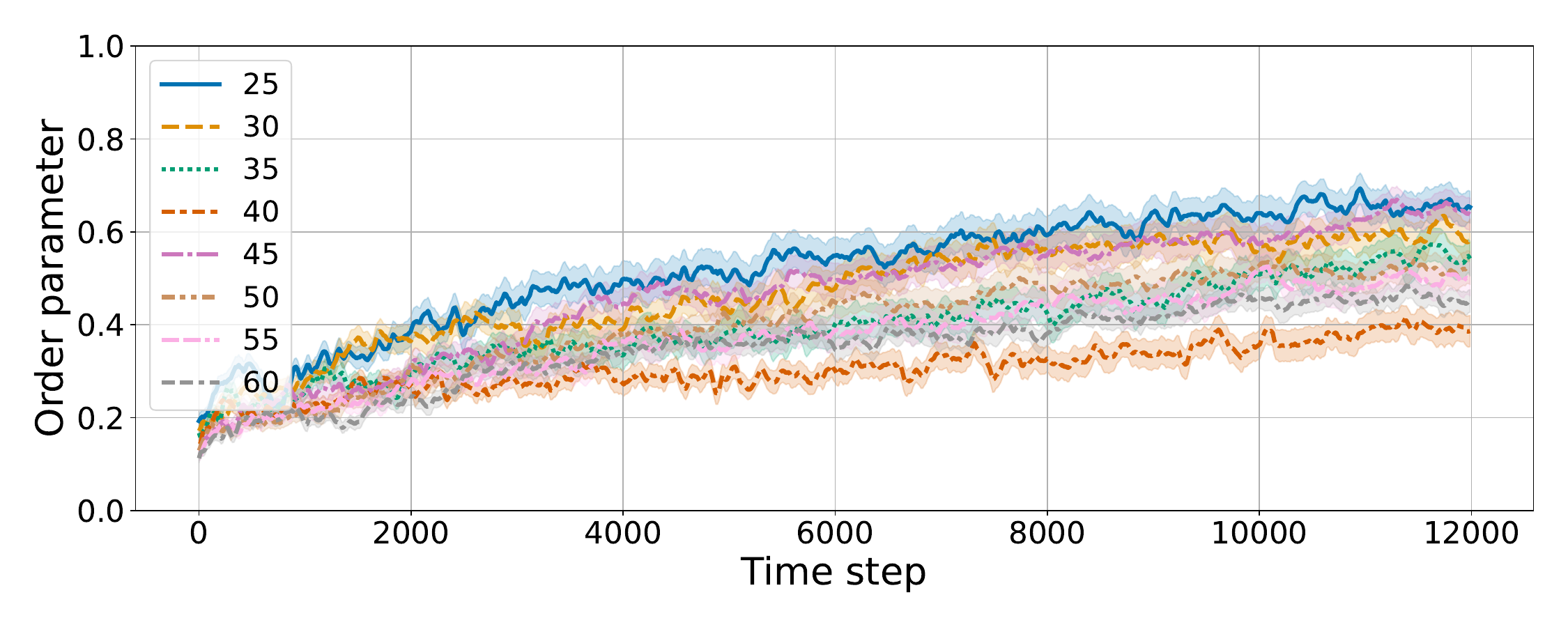}
        \caption{20\% Faulty robots}
        \label{fig:20_faulty}
    \end{subfigure}
    \caption{Evolution of order for the AA-V model across different swarm sizes with varying proportions of faulty robots.}
    \label{fig:multi_faulty_robots}
\end{figure*}

To illustrate, we experiment with different percentages of robots that are stuck in place, in swarm sizes 25--60 (\Cref{fig:multi_faulty_robots}). The swarm uses the AA-V model.  In all sub-figures, performance is measured by order, excluding all faulty robots from the analysis.  The figure shows that as the proportion of failed robots increases,
and the swarms are slower to reach an ordered state, if they reach it at all.
Larger swarms appear to degrade more easily, given a percentage of failing robots \citep{bjerknes13, shefi24}.

We discuss the principles of visual detection of such faulty neighbors in~\cref{sec:robots_fail_perception_not_perfect}. We then discuss intermittent locomotion as a way to facilitate such detection in practice (\cref{sec:aapg}). Finally, we evaluate the different techniques in extensive experiments (\cref{sec:results}).

\subsection{Visual Detection of Faulty Robots}
\label{sec:robots_fail_perception_not_perfect}
For each focal robot $i\in \mathcal{A}$, the neighbor detection process provides a set of neighbors $\mathcal{D}_i^t$.
Both AA and AA-V models do not consider the neighbor's velocity in computing $i$'s motion vector $F_i^t$ (\cref{eq:avoid-attract} and \cref{eq:rjvector}).
They therefore cannot detect a robot that is too slow, and so cannot isolate its effects on the individual motion computation. We examine this formally below, from the perspective of the observing focal robot.

\paragraph{Visual Perception of Neighbor Motion}
Each of the detected $j\in \mathcal{D}_i^t$ is represented by two position vectors in the focal robot's frame of reference, $\Lt{t}$ and $\Rt{t}$, from which a distance-corrected neighbor position vector $\Ct{t}$ is computed.
We assume that once a neighbor is visible, it is visually tracked until it is no longer visible (i.e., becomes occluded or lies outside $r_{sense}$).
Thus, as long as a specific neighbor $j$ is detected, the $\Lt{},\Rt{}$ remain associated with it, despite being recomputed at each new step. This does not imply distinguishing robots in $\mathcal{A}$, or handling of intermittent discontinuities in tracking: if robot $j$ is visible at time $t$, but then becomes undetected, and returns to be visible at time $t+k$, it is not assumed that the vectors $\Lt{t},\Rt{t}$ and $\Lt{t+k},\Rt{t+k}$  will be associated with it.

Given vectors at two times $t$ and $t+1$ in \emph{the same coordinate system}, and \emph{for the same neighbor} one may easily deduce that movement occurred, if there is a change in one or both of these vectors, i.e., if $\Lt{t+1}\neq\Lt{t}$, or $\Rt{t+1}\neq \Rt{t}$.

Such changes may imply a change in the position of the neighbor (i.e., $\Ct{t+1}\neq \Ct{t}$).
Specifically, the instantaneous position change vector is
\begin{equation}
	\vec{\delta_U} = \Ct{t+1}-\Ct{t}.
\end{equation}
Converting $\vec{\delta_U}$ from polar to rectangular coordinates, the instantaneous \textit{linear velocity} is given as
\begin{equation}
	\delta_U = \left\|\vec{\delta_U}\right\|.
\end{equation}

This is not the only way in which the motion of a neighbor may be perceived. The neighbor may change its orientation (i.e., turn in place), while maintaining its distance and bearing. In this case $\Ct{t+1}=\Ct{t}$, but $\Lt{t+1}\neq\Lt{t}$ and $\Rt{t+1}\neq \Rt{t}$.
To detect such movement, the focal robot must compute the \emph{rotational velocity} of the neighbor by noting a change in the angular width of the robot from time $t$ to time $t+1$.
Given the angle $\theta^t$ between $\Lt{t}, \Rt{t}$  (and respectively, $\theta^{t+1}$), the change in the angular width of the neighbor gives the perceived rotational change.
Practically, we prefer measuring the relative rotational change:
\begin{equation}
	\delta_\theta = \frac{\theta^{t+1}-\theta^{t}}{\theta^{t+1}}.
\end{equation}

\paragraph{Perception of Faulty Neighbors}
Using $\delta_U$ and $\delta_\theta$, the focal robot can detect if a neighbor moves too slowly. Given thresholds $U_{min}, \Theta_{min}$, we can define a failing robot:
\begin{definition}
	A neighbor robot $j\in\mathcal{D}_i$ is considered faulty by focal robot $i\in\mathcal{A}$ if its associated $|\delta_U|\leq U_{min}$ or $|\delta_\theta|\leq\Theta_{min}$ \emph{for the entire period it is visible, during the pause state}.
	\label{def:failed_robot}
\end{definition}

Examining all neighbors $j\in \mathcal{D}_i^t$ through the lens of~\cref{def:failed_robot} partitions $\mathcal{D}_i^t$ into two non-overlapping subsets $\mathcal{N}_i^t$ (nominal robots, for whom~\cref{def:failed_robot} did not hold), and $\mathcal{S}_i^t$ (robots considered to be faulty by~\cref{def:failed_robot}). The neighbor selection stage makes the distinction between these, so that later motion computation can treat them separately (as discussed in~\cref{sec:interacting}).

However, there is a key challenge with respect to the use of $\delta_U$ and $\delta_\theta$ in the context of~\cref{def:failed_robot}. Both are computed \textit{relative to the focal robot's frame of reference}; they measure \emph{relative} movement.
Critically, if the focal robot is moving while tracking the neighbor, then $\delta_U$ and $\delta_\theta$ reflect the combination of the neighbor motions, and the focal robot's ego-motion (which shifts the polar coordinates' origin point).

This can lead to incorrectly partitioning neighbors into the sets $\mathcal{N}_i$ and $\mathcal{S}_i$.
For example, a neighbor whose movement exactly parallels (or complements, as the case may be) that of the focal robot will appear to be standing still.
Alternatively, a faulty neighbor that is stuck in place will appear to move away, because the focal robot is moving; or it may appear to be rotating around its center (though the perceived rotation is due to the focal robot's movement).

In principle, if the focal robot knows its own ego-motion, it can nullify those effects on the relative velocity measurements.
This can be done by computing the expected ego-motion effects, and subtracting them from the observed relative velocities $\delta_U, \delta_\theta$.

Unfortunately, excluding the idealized sensing of the AA model, the robot's ego-motion is not inherently known to it.
We have assumed only a visual sensor and some basic capabilities for neighbor recognition and tracking.
This does not mean that ego-motion is known.
Indeed, the interaction of the focal robot and the environment is fundamentally open-loop.
No matter how good the actuation is in translating control signals to target movement, there are exogenous forces in the environment that may act on the robot and move it without its control~\citep{borenstein96}, e.g., wheel slippage or neighbors
physically bumping and pushing on the robot.
Note also that the focal robot itself may be faulty: it may be sending control commands that are not being carried out.
The actual ego-motion must be sensed and estimated through the visual perception processes.

\subsection{Pause-and-Go Locomotion}
\label{sec:aapg}
We observe that while the focal robot does not know its ego-motion, it can take actions to nullify its effects in its own perception, \emph{by stopping}.
If it pauses its movement, its own motion is ideally zero, and so any perceived neighbor velocities would allow distinguishing nominal and faulty robots. Note that this is an approximation: even if the robot
stops in place, it is still subject to movements caused by exogenous forces such as robots bumping into it.
However, if it is surrounded by nominal robots, they will be attempting to avoid bumping into it, and so the approximated zero ego-motion may be correct most of the time.

Building on this observation, we are inspired by the individual asynchronous intermittent locomotion in locust nymphs marching in cohesive bands of millions.  Each locust alternates between a movement state
and a stopping state~\citep{aidan24}.
The natural reason for this intermittent motion is not clear.
However, applying it in robot swarms can have great benefits.

We introduce AAPG-V, a pause-and-go (P\&G) collective motion model, extending the AA-V model introduced earlier.  In this model, every focal robot alternates between moving (\textit{go} phase) and stopping (\textit{pause} phase). During the pause phases, the focal robot observes its neighbors and detects faulty robots that meet the criteria of~\cref{def:failed_robot} for the entire duration of the phase. This allows it to select neighbors for the $\mathcal{N}_i$ and $\mathcal{S}_i$ sets. During the go phase, the robot moves based on motion computation that respects the differences between these sets.
This allows nominal neighbors to detect that the focal robot itself is faulty, even if it does not know this (i.e., it sends actuation commands that are ineffective).

\subsubsection{AAPG-V Overview}
\label{sec:Pause_and_Go_for_Fault_Detection}
We begin with a brief overview of the AAPG-V model (details in the next sections).
It builds on the AA-V model, relying on its visual sensor and assumed capacity to detect neighbors in occluded settings, generating the set $\mathcal{D}_i$ for every $i\in \mathcal{A}$, at all times.
It adds the tracking capabilities discussed above, so as to estimate the velocities of neighbors.
These are used to partition neighbors that are nominal (the $\mathcal{N}_i$ set) and faulty (the $\mathcal{S}_i$ set).
Each robot alternates between two behavioral states:
\begin{itemize}
	\item When in the \textit{pause} state, the robot stops its own motion (no position change or rotation), and observes its neighbors. At the beginning of this state, the sets $\mathcal{N}_i, \mathcal{S}_i$ are reset, and are then computed from observations during the pause.
	\item When in the \textit{go} state, robots compute a motion vector $\vec{G_i^t}$ that addresses faulty neighbors differently than nominal robots. Only the set $\mathcal{N}_i$ is updated when in this state.
\end{itemize}
\cref{eq:mdmc_p_and_g} presents this formally:
\begin{equation}
	\vec{F_i^t} =
	\begin{cases}
		\langle 0,0\rangle & \text{state is \emph{pause}, see~\cref{sec:misclassification_faulty_robots}} \\
		\vec{G_i^t} & \text{state is \emph{go}, see \cref{sec:interacting}}.
	\end{cases}
	\label{eq:mdmc_p_and_g}
\end{equation}
P\&G locomotion is parameterized by two separate intervals, $\underrightarrow{P}=[P_{min}, P_{max})$ and $\underrightarrow{G}=[G_{min},G_{max})$, defining minimum and maximum durations for the go and pause states.  When transitioning from a pause state to a go state, the robot $i$ randomly selects a go duration $g_i\in \underrightarrow{G}$. When its movement duration ends, the robot transitions from a go state to a pause state, analogously selecting a pause duration $p_i\in\underrightarrow{P}$. This repeats independently and asynchronously for every robot $i\in \mathcal{A}$. As a result, at any time $t$, some robots are moving, while others are paused.

\subsubsection{Pausing to Observe}
\label{sec:misclassification_faulty_robots}
Upon transitioning to the pause state, the robot first randomly picks a pause duration $p_i$ from the pause interval $\underrightarrow{P}$. It sets its movement motion vector $F_i$ to $(0,0)$ (\cref{eq:mdmc_p_and_g}),
stopping all ego-motion.  It resets the sets $\mathcal{N}_i$ and $\mathcal{S}_i$ to be empty.

Let us denote the time of the transition into the pause state as $t_0$. During the pausing interval $[t_0,t_0+p_i]$, the robot continues to observe its surroundings and detect neighbors. For any time
$t\in [t_0,t_0+p_i]$, the neighbor detection stage provides the set of detected neighbors $\mathcal{D}_i^t$.

Starting at time $t_0+1$, the neighbor selection stage examines pairs of detected neighbor sets $\mathcal{D}_i^{t-1}, \mathcal{D}_i^{t}$.  For each robot $j$ appearing in these, there are several possibilities:
\begin{enumerate}
\item $j\in \mathcal{D}_i^{t-1}\wedge j\in \mathcal{D}_i^{t}$. $j$ is a neighbor that has been tracked across two consecutive time-steps. There are two cases:
\begin{enumerate}
	\item $j\notin \mathcal{N}_i$, and $j\notin \mathcal{S}_i$. This is a new neighbor to be classified as either nominal or faulty. Add to $\mathcal{S}_i$ if~\cref{def:failed_robot} applies. \label{special-pause-case}
	 Otherwise, add to $\mathcal{N}_i$.
	\item $j\in \mathcal{S}_i$. This is a neighbor who is thought to be faulty. If~\cref{def:failed_robot} no longer applies to it (it passes the movement thresholds for time $t$), then remove it from $\mathcal{S}_i$ and add it to $\mathcal{N}_i$.
\end{enumerate}
\item $j\in \mathcal{D}_i^{t-1}\wedge j\notin \mathcal{D}_i^t$. This is a neighbor that is no longer detected. Even if it reappears in the future, it will not be recognized as the same neighbor. Thus, it can be safely removed from consideration. If it is in $\mathcal{N}_i$ or $\mathcal{S}_i$, it is removed from them.

\item $j\notin \mathcal{D}_i^{t-1}\wedge j\in \mathcal{D}_i^t$. This is a neighbor that has just been detected at time $t$. It has either moved into the sensing range or was perhaps occluded until now by a different robot that has moved. In either case, we keep tracking and do not yet add to any set.
\end{enumerate}
In all other cases, there is nothing to be done; the sets $\mathcal{N}_i, \mathcal{S}_i$ remain unchanged.
Once the pausing duration ends, the sets $\mathcal{N}_i, \mathcal{S}_i$ are available for use (and updates) in the go phase. We note that \cref{special-pause-case} reflects an interpretation that a robot is considered faulty if \cref{def:failed_robot} applies to it for the continuous time it is visible. Alternatively, it may be considered faulty if it \cref{def:failed_robot} applies to the entire duration of the pause.

\begin{figure}[htbp]
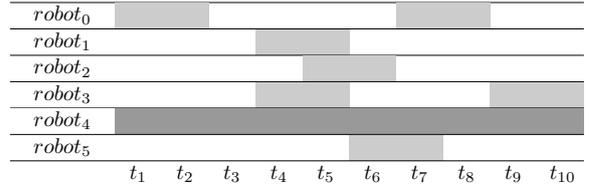

	\centering
	\scalebox{0.8}{\centering
		\begin{tabular}{p{1.3cm}*{10}{p{0.35cm}}}

			\hline
			\centering $robot_0$ & \cellcolor{gray!40} &\cellcolor{gray!40}& & & & &\cellcolor{gray!40} &\cellcolor{gray!40} & \\
			\hline
			\centering $robot_1$ & &  & &\cellcolor{gray!40} &\cellcolor{gray!40}  & & & & &\\
			\hline
			\centering $robot_2$ & & &  & &\cellcolor{gray!40} &\cellcolor{gray!40} & & & &\\
			\hline
			\centering $robot_3$ & & & & \cellcolor{gray!40} &\cellcolor{gray!40} & & & &\cellcolor{gray!40} &\cellcolor{gray!40}\\
			\hline
			\rowcolor{black!40} \centering $robot_4$ \cellcolor{white} & & & & & & & & & &\\
			\hline
			\centering $robot_5$ & & & & & &\cellcolor{gray!40} &\cellcolor{gray!40} & & &\\
			\hline
			&\centering $t_1$ &\centering $t_2$ &\centering $t_3$ &\centering $t_4$ &\centering $t_5$ &\centering $t_6$ &\centering $t_7$ &\centering $t_8$ &\centering $t_9$ &\centering $t_{10}$\\\end{tabular}
	}
	\caption{Example timeline. Each row represents the states of a swarm robot over time (horizontal axis). Grayed cells indicate pause states; white cells indicate go-states. Robot $robot_4$ is faulty (dark gray), and would be correctly identified by all. In contrast, $robot_1$ and $robot_3$ will incorrectly consider each other to be faulty, since their pause durations overlap in $t_4$--$t_5$. }
	\label{fig:timeline}
\end{figure}

While pausing gives the focal robot a chance to observe and classify its neighbors, it also elevates the risk that it would be classified as faulty by others, observing it does not move.
To illustrate,~\cref{fig:timeline} shows how two robots that pause at the same time for the same duration may classify each other as faulty, as no motion is observed.

\citet{shefi24} raises the idea of \emph{early termination of the pause state}, allowing robots to cut short the duration of the pause state before $p_i$ is reached if some conditions are met.
While they report on improvements achieved using such rules, it was in the context of the AA model (idealized sensing, no occlusions).
When we duplicated these rules for use in the AAPG-V model, no substantial benefits that would justify the added complexity were demonstrated empirically.
We therefore do not investigate this further.

\subsubsection{Going (Moving)}
\label{sec:interacting}
When transitioning to the go state, the robot first randomly picks a go duration $g_i$ from the go interval $\underrightarrow{G}$. We again denote the time of the transition as $t_0$.  At any time $t\in [t_0,t_0+g_i]$,
the robot computes the motion vector $G_i^t$ (used in~\cref{eq:mdmc_p_and_g}) according to~\cref{eq:aapg}:
\begin{equation}
	\vec{G_i^t} = \sum_{j\in \mathcal{N}_i^t}\vec{f^t_{ij}} + \sum_{j\in \mathcal{S}_i^t}\vec{s^t_{ij}},
	\label{eq:aapg}
\end{equation}
where $\vec{f^t_{ij}}$ denotes the contribution from nominal (healthy) neighbors (\cref{eq:rjvector}), and $\vec{s^t_{ij}}$ denotes interactions with faulty (stuck) neighbors (described below).  While it moves, new neighbors appearing in $\mathcal{D}_i^t$ that are not in $\mathcal{S}_i$ or $\mathcal{N}_i$ are added to $\mathcal{N}_i$ (i.e., they are considered nominal, as the perceived motion cannot be verified).

The basic version of $\vec{s^t_{ij}}$ is given in~\cref{eq:rjvector-avoid}. It uses the Heaviside step function $\mathbb{H}$ to distinguish faulty neighbors that are farther than the desired distance $r_d$ (and are thus to be ignored) from those that are closer than $r_d$ (and should therefore be avoided).
\begin{equation}
	\vec{s_{ij}} = \left( \frac{r_{ij} - r_d}{r_{ij}^2} \cdot \mathbb{H}(r_d - r_{ij}),\, \angle \beta_{ij} \right)
	\label{eq:rjvector-avoid}
\end{equation}

\begin{figure}[htbp]
	\centering
	\includegraphics[width=\columnwidth]{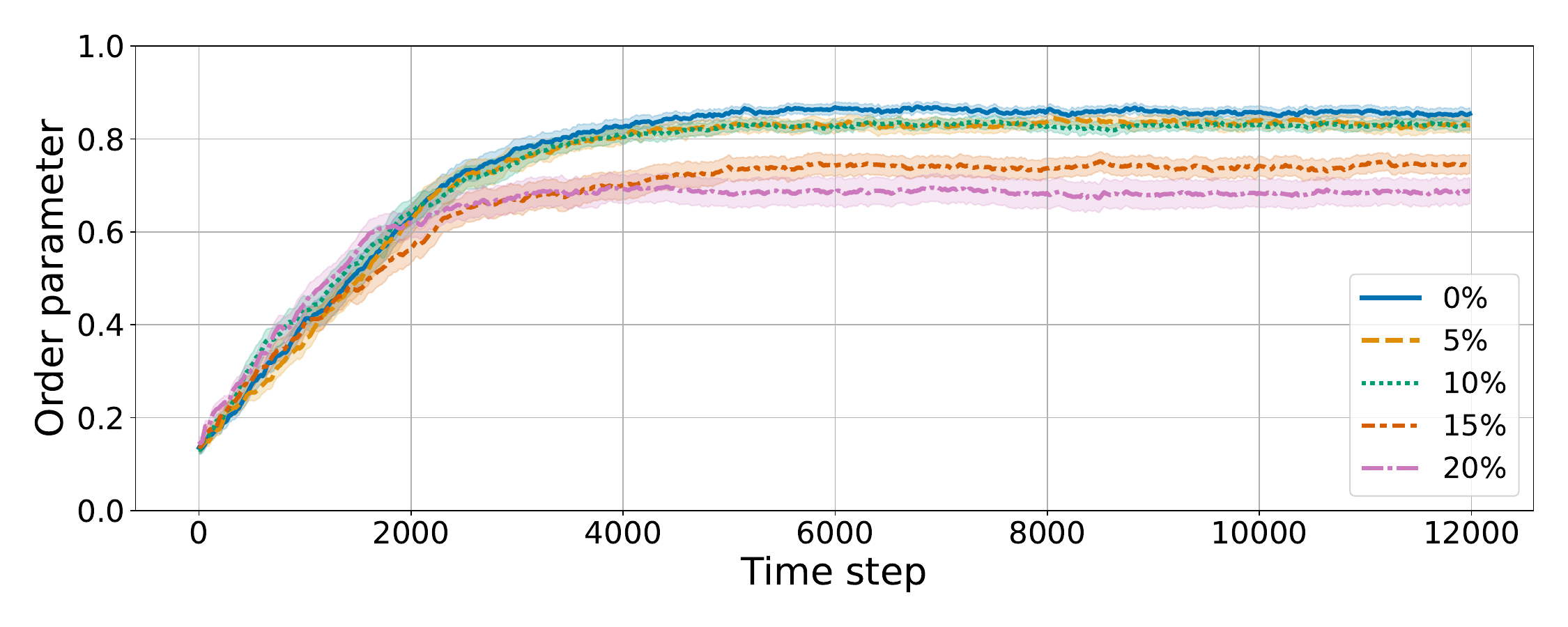}
	\caption{AAPG-V with varying proportions of faulty robots (out of 40), using the avoid method (\cref{eq:rjvector-avoid}) under perfect fault detection.}

	\label{fig:perfect_detection}
\end{figure}

\Cref{fig:perfect_detection} illustrates swarm order for a population of 40 robots under varying proportions of faulty robots, \emph{with perfect detection}. We set $\underrightarrow{P}=[6,7),\underrightarrow{G}=[11,20)$ (empirically determined; see~\cref{sec:results}).
In all cases, the swarm converges to a coherent state, though the steady-state order decreases as the proportion of faulty robots increases. These results demonstrate the viability of the approach.

\begin{figure}[htbp]
	\centering
		\includegraphics[width=0.8\columnwidth]{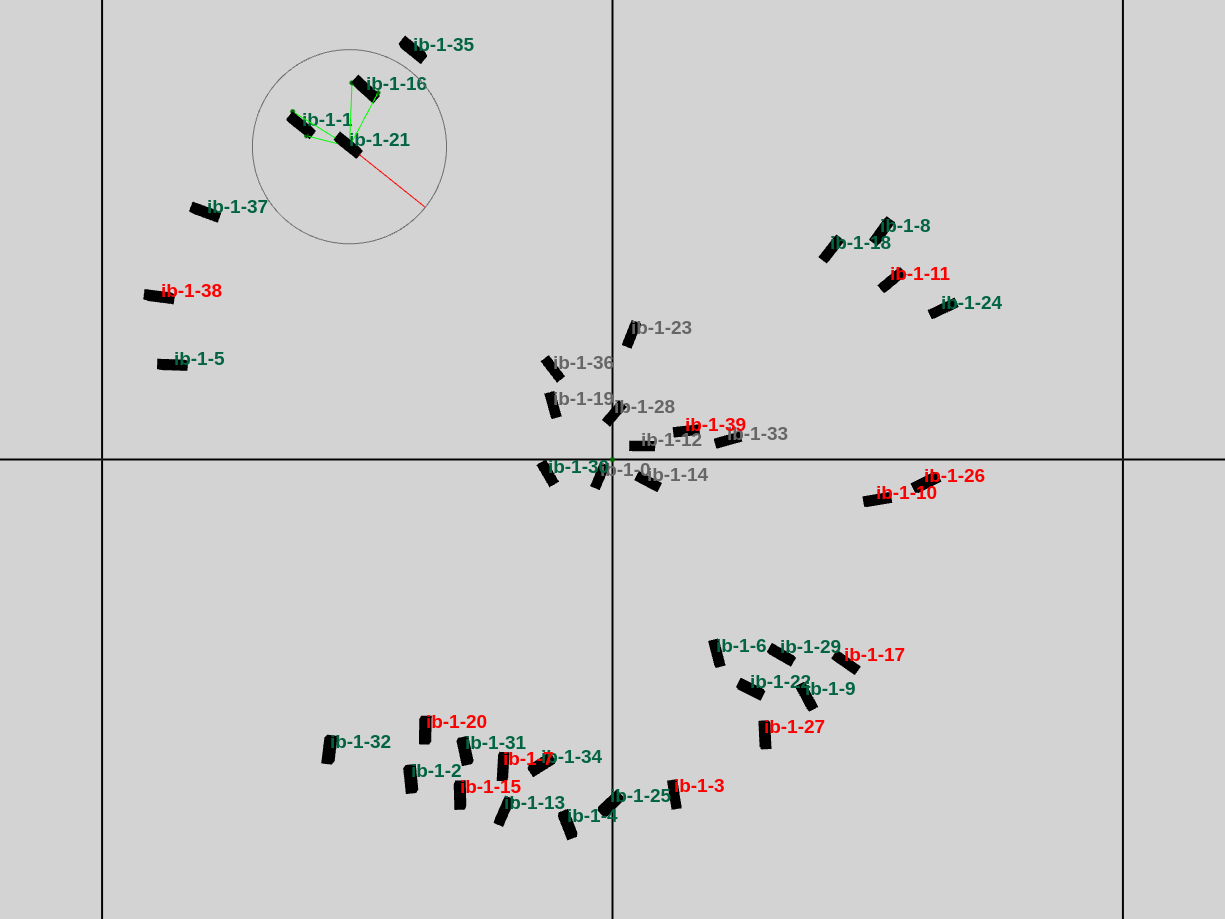}

	\caption{Simulation screenshot showing fragmentation of a 40-robot swarm, where 20\% (8) robots are faulty (stuck in place). Label colors indicate robot states: \textit{green} for moving, \textit{red} for paused, and \textit{gray} for failed.
	}
	\label{fig:avoid-screenshots}
\end{figure}

As~\cref{fig:timeline} demonstrates, detection under the P\&G scheme can be imperfect, as two nominal robots that pause simultaneously ($robot_1, robot_3$)  may erroneously consider each other to be faulty.
The result of avoiding robots that have been incorrectly classified as faulty leads to \emph{fragmentation} in the swarm, forming multiple ordered sub-swarms moving in different directions (\cref{fig:avoid-screenshots}).

\paragraph{Managing Misclassification}
Given that misclassification of faulty robots occurs, we require more cautious approaches that mitigate the risk to the order of the motion in this case.  For instance, rather than avoiding faulty members, we may use the regular avoid-attract vector $\vec{f^t_{ij}}$, but with a much smaller gain (we use 1 for nominal robots).  Empirically, we found this to be ineffective.

Instead, we apply a stochastic policy, where each suspected neighbor is randomly treated at every time step as either faulty (avoid) or nominal (standard AA interaction).
At every step $t$, and for each neighbor $j\in \mathcal{S}_i$, the motion computation decides (with probability $p$) to treat $j$ as faulty, using~\cref{eq:rjvector-avoid}. With probability (1-$p$), it treats it as nominal per~\cref{eq:rjvector}:
\begin{equation}
\vec{s_{ij}^t} =
\begin{cases}
\left( \tfrac{r_{ij}-r_d}{r_{ij}^2}\!\cdot\!\mathbb{H}(r_d-r_{ij}),\,\angle\beta_{ij} \right)
 & \text{w/ prob.} \;  p, \\
  \left(\frac{r_{ij} - r_d}{r_{ij}^2}\,,\,\qquad\qquad\quad \angle\beta_{ij}\right) & \text{otherwise}.
  \label{eq:avoid-halftime}
\end{cases}
\end{equation}
This balances the impact of actual and misclassified faulty neighbors on the swarm.

\begin{figure}[htbp]
    \centering

        \includegraphics[width=\columnwidth]{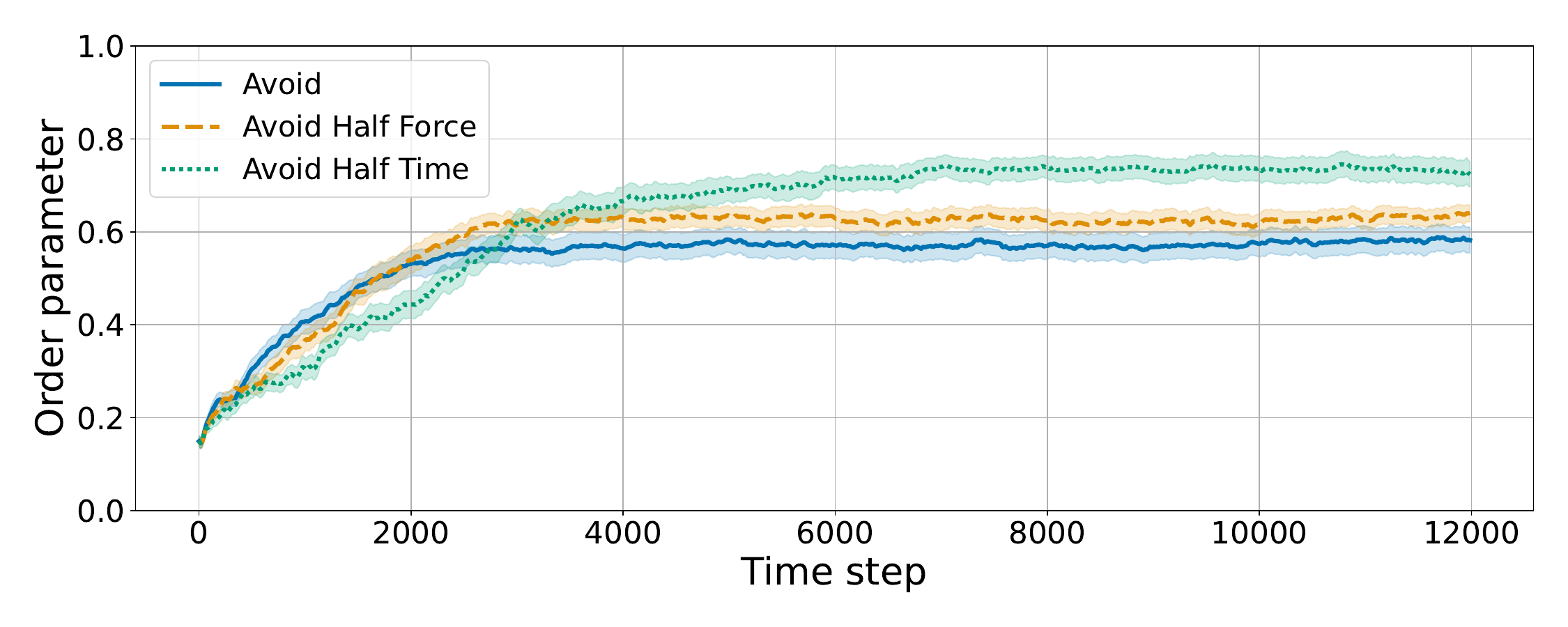}

    \caption{Comparison of alternative interaction rules in the AAPG-V model with 40 robots, of which 20\% (8) are faulty. Detection of faulty robots takes place during a pause (realistic). }
    \label{fig:interaction_rules}
\end{figure}

In \cref{fig:interaction_rules}, we compare the different interaction strategies under realistic fault detection settings, rather than  the idealized perfect detection in~\cref{fig:perfect_detection}.
The \emph{Avoid} method uses~\cref{eq:rjvector-avoid}. It ignores faulty neighbors that are farther than the desired distance $r_d$, and avoids those that are closer than $r_d$. 
The \emph{Avoid Half Force} method uses \cref{eq:rjvector} for faulty members, with a gain of 0.5. The \emph{Avoid Half Time} method is stochastic, using~\cref{eq:avoid-halftime} with probability $p=0.5$.
In these experiments, we assume that the thresholds $U_{min}, \Theta_{min}$ are set to 0, i.e., a robot is considered faulty if it appears to be completely motionless. In~\cref{ch:Discussion} we revisit this assumption.

\begin{figure}[htbp]
	\centering

	\includegraphics[width=\columnwidth]{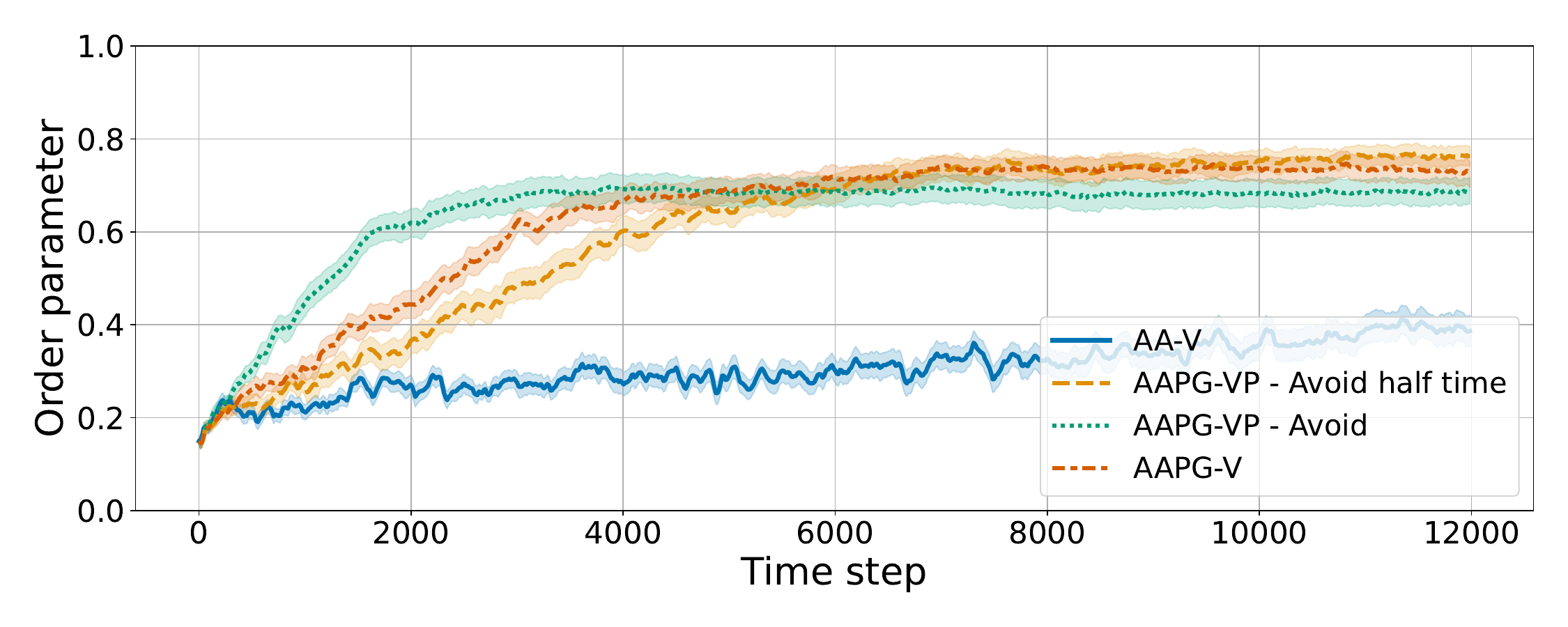}

	\caption{Comparison of alternative interaction rules in the AAPG-V model with 40 robots, of which 20\% (8) are faulty.
		The AAPG-V model (red dashed line) marks the results from the AAPG-V model with realistic misclassification.
		The others show results under idealized fault detection conditions.
		 }
	\label{fig:interaction_rules_perfect}
\end{figure}

\emph{Avoid Half-Time} is clearly superior to the other methods in the more realistic settings, where misclassification cannot be avoided.
However, we should also compare it to a hypothetical variant, which uses idealized detection of failure.
This allows a better understanding
of the unavoidable impact of misclassification on the order of the swarm.
\Cref{fig:interaction_rules_perfect} presents this comparison.
For perspective, the results from an idealized AAPG-V \emph{Avoid} method, and the AA-V model are also shown.
The realistic AAPG-V model is only slightly worse than its theoretical counterpart (using idealized fault detection).
It outperforms the others. Note that the AA-V model is shown for comparison only. Even with idealized detection, it has no way
to interact differently with faulty robots, and so fails to generate collective motion.

\subsection{AAPG-V: Results}
\label{sec:results}

The choice of pause and go interval parameters $\underrightarrow{G},\underrightarrow{P}$ directly impacts classification accuracy, but also the order of the swarm, its potential fragmentation, and speed.
Longer pause phases provide more time to observe others, but simultaneously increase the chance of being mistaken for faulty. Longer go phases thus allow robots to move farther and cover more ground, but at a higher risk of disruption, due to not recognizing faulty robots.

We do not yet have a theory of how interval parameter choice impacts the performance criteria of the swarm.
We therefore explore their setting empirically. \cref{app:pgsensitive} provides the results of extensive experiments measuring both the misclassification rate when different interval durations are used, as well as the order achieved.
We observe qualitative general trends: shorter pause durations coupled with longer go pauses tend to produce
higher rates of misclassification, but also, paradoxically, higher order.
Based on the results, we chose $\underrightarrow{P}=[6,7),\underrightarrow{G}=[11,20)$. We leave an in-depth analysis of the relationship between order and misclassification rate under different settings for future work.

\paragraph{Collective Motion Order}
We compare the AA-V model with our proposed fault-tolerant AAPG-V model using the stochastic \emph{half-time avoidance} method (\cref{eq:avoid-halftime}, with probability $p=0.5$).

\begin{figure*}[htp]
    \centering
\begin{subfigure}{0.33\textwidth}
        \centering
        \includegraphics[width=1.05\columnwidth]{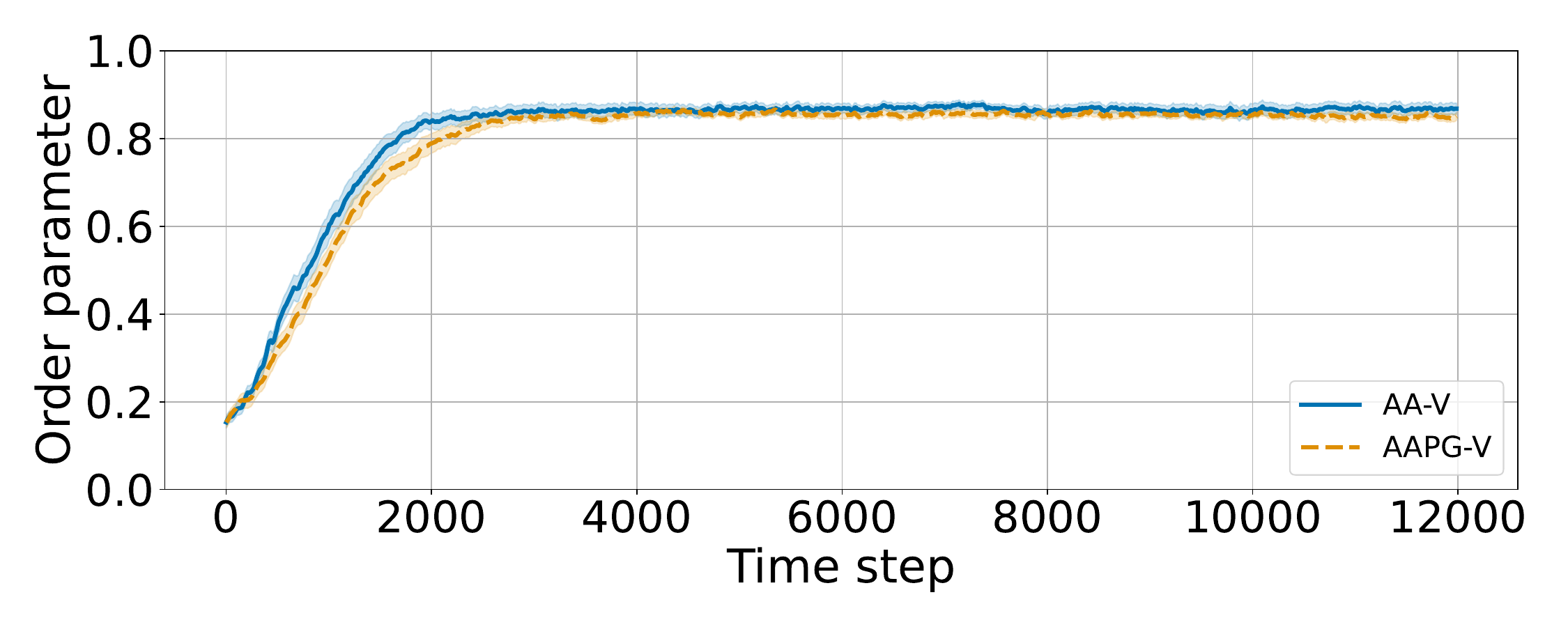}
        \caption{25 robots, 0\% faulty.}
    \end{subfigure}
    \hfill
    \begin{subfigure}{0.33\textwidth}
        \centering
        \includegraphics[width=1.05\columnwidth]{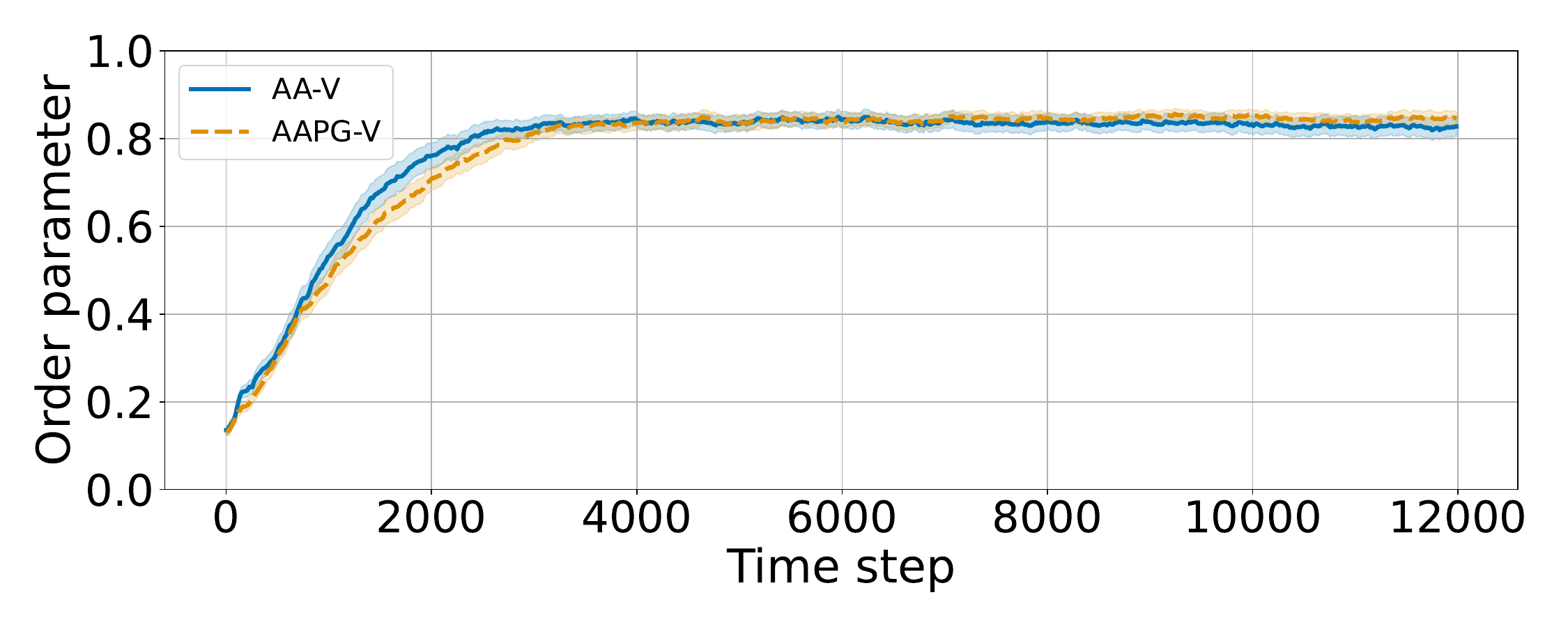}
        \caption{40 robots, 0\% faulty.}
    \end{subfigure}
    \hfill
    \begin{subfigure}{0.32\textwidth}
        \centering
        \includegraphics[width=1.05\columnwidth]{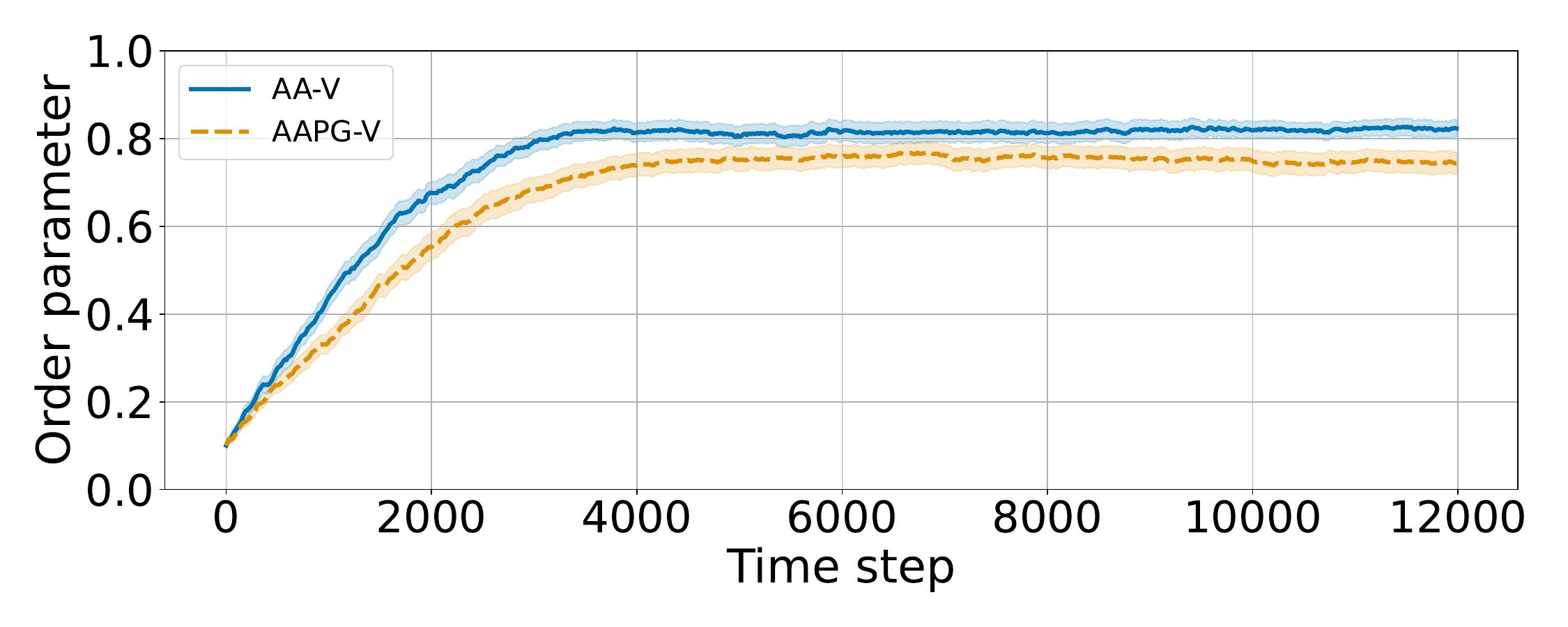}
        \caption{60 robots, 0\% faulty.}
    \end{subfigure}

    \vspace{0.3cm}

\begin{subfigure}{0.33\textwidth}
        \centering
        \includegraphics[width=1.05\columnwidth]{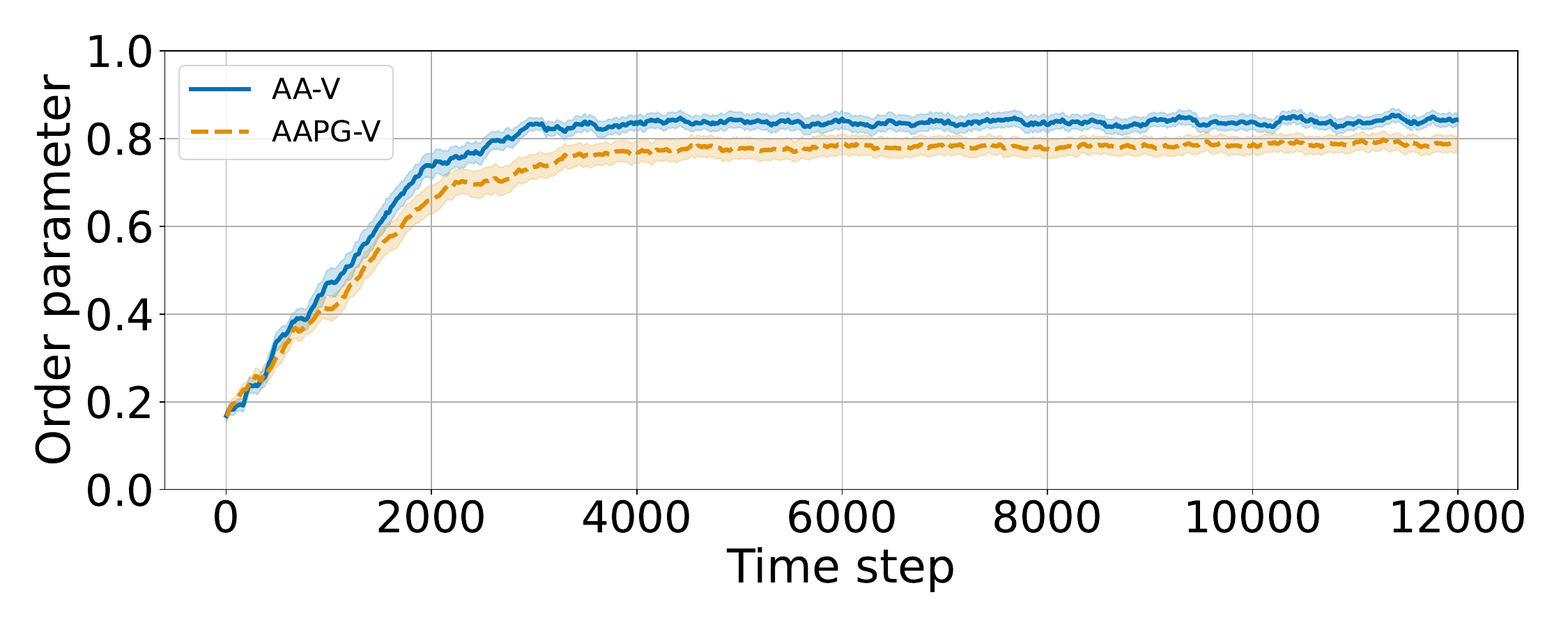}
        \caption{25 robots, 10\% faulty.}
    \end{subfigure}
    \hfill
    \begin{subfigure}{0.33\textwidth}
        \centering
        \includegraphics[width=1.05\columnwidth]{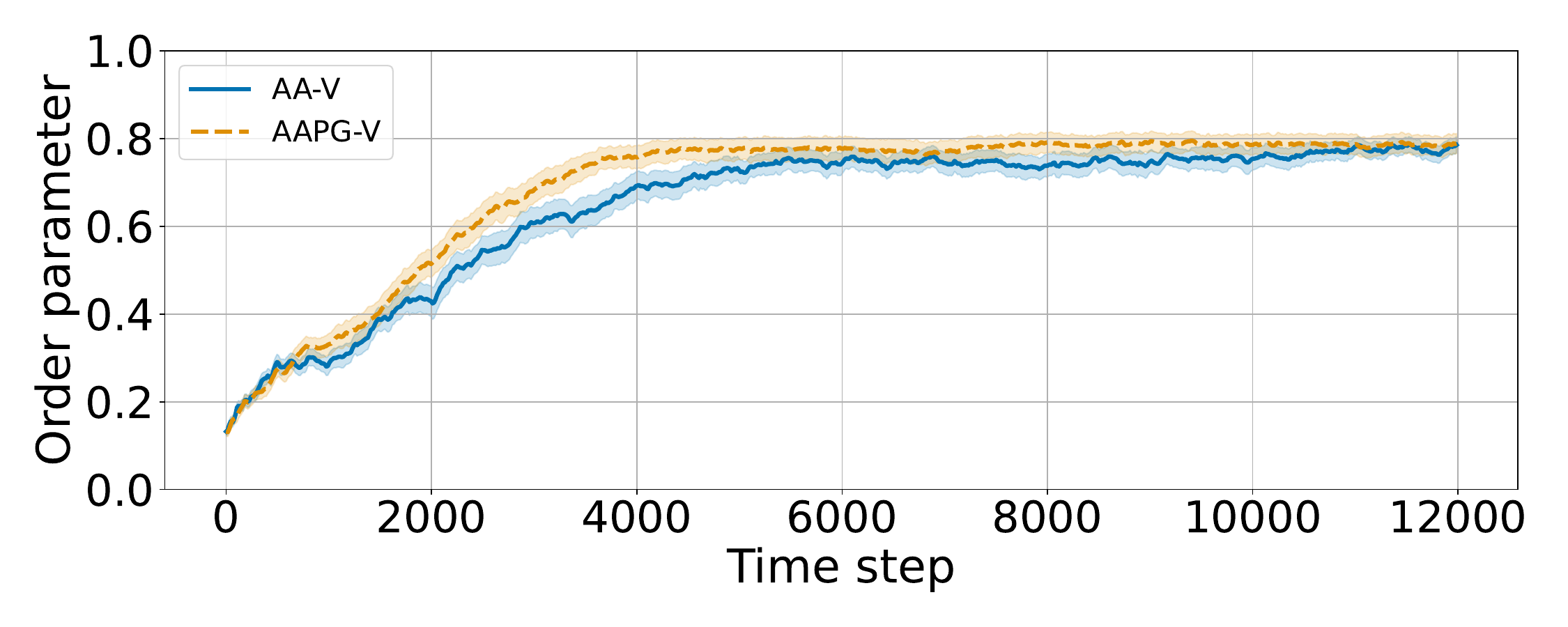}
        \caption{40 robots, 10\% faulty.}
    \end{subfigure}
    \hfill
    \begin{subfigure}{0.32\textwidth}
        \centering
        \includegraphics[width=1.05\columnwidth]{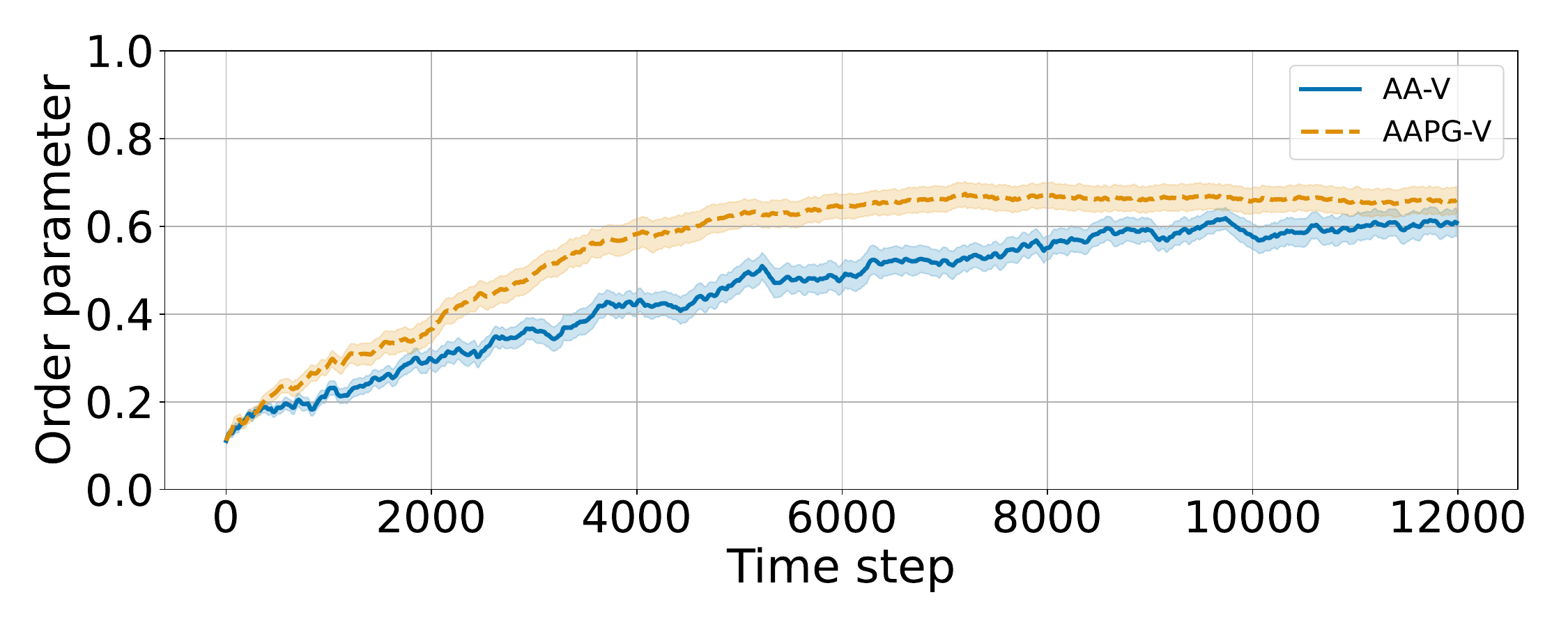}
        \caption{60 robots, 10\% faulty.}
    \end{subfigure}

    \vspace{0.3cm}

\begin{subfigure}{0.33\textwidth}
        \centering
        \includegraphics[width=1.05\columnwidth]{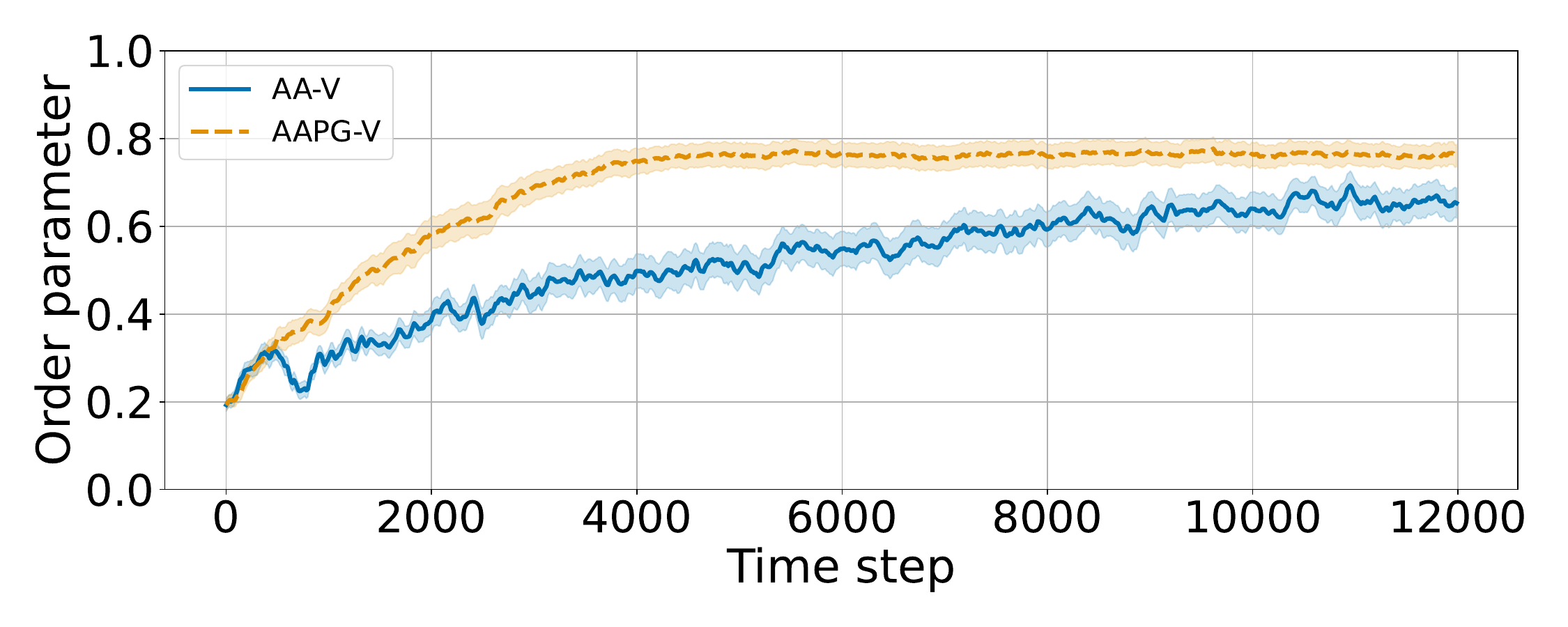}
        \caption{25 robots, 20\% faulty.}
    \end{subfigure}
    \hfill
    \begin{subfigure}{0.33\textwidth}
        \centering
        \includegraphics[width=1.05\columnwidth]{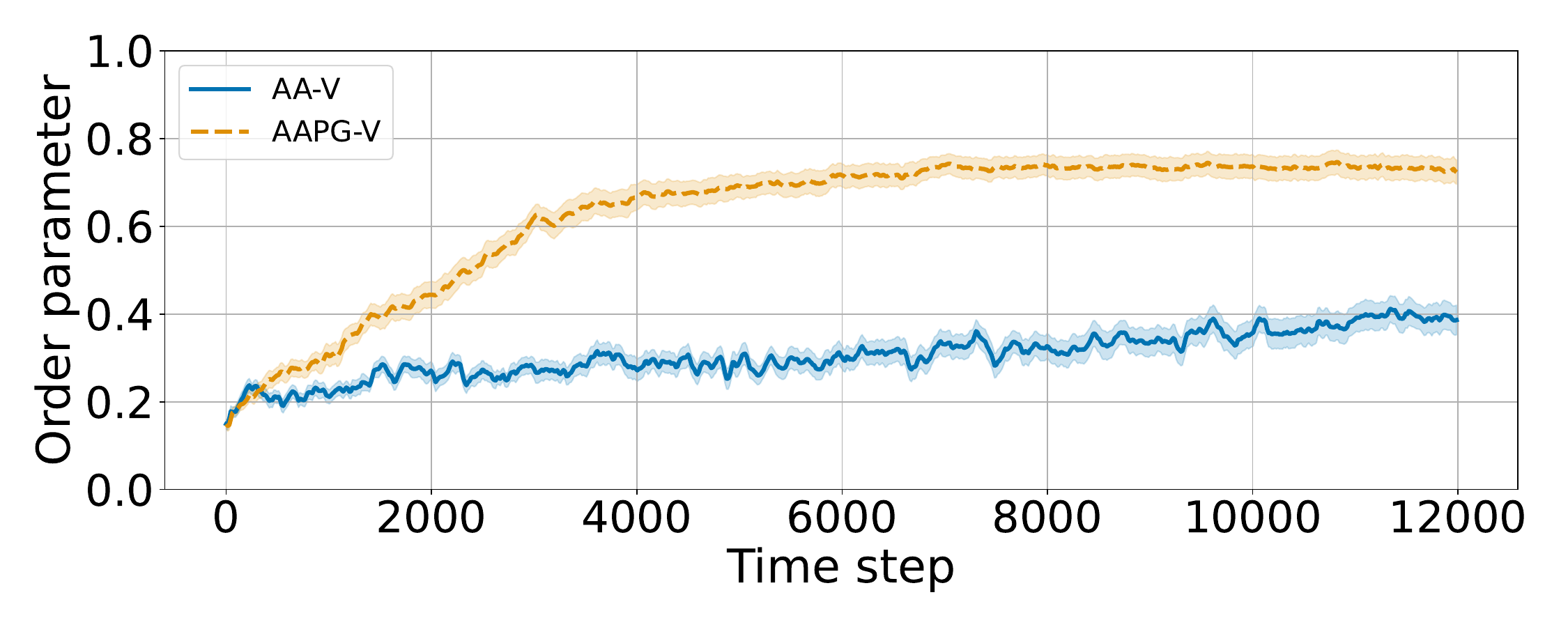}
        \caption{40 robots, 20\% faulty.}
    \end{subfigure}
    \hfill
    \begin{subfigure}{0.32\textwidth}
        \centering
        \includegraphics[width=1.05\columnwidth]{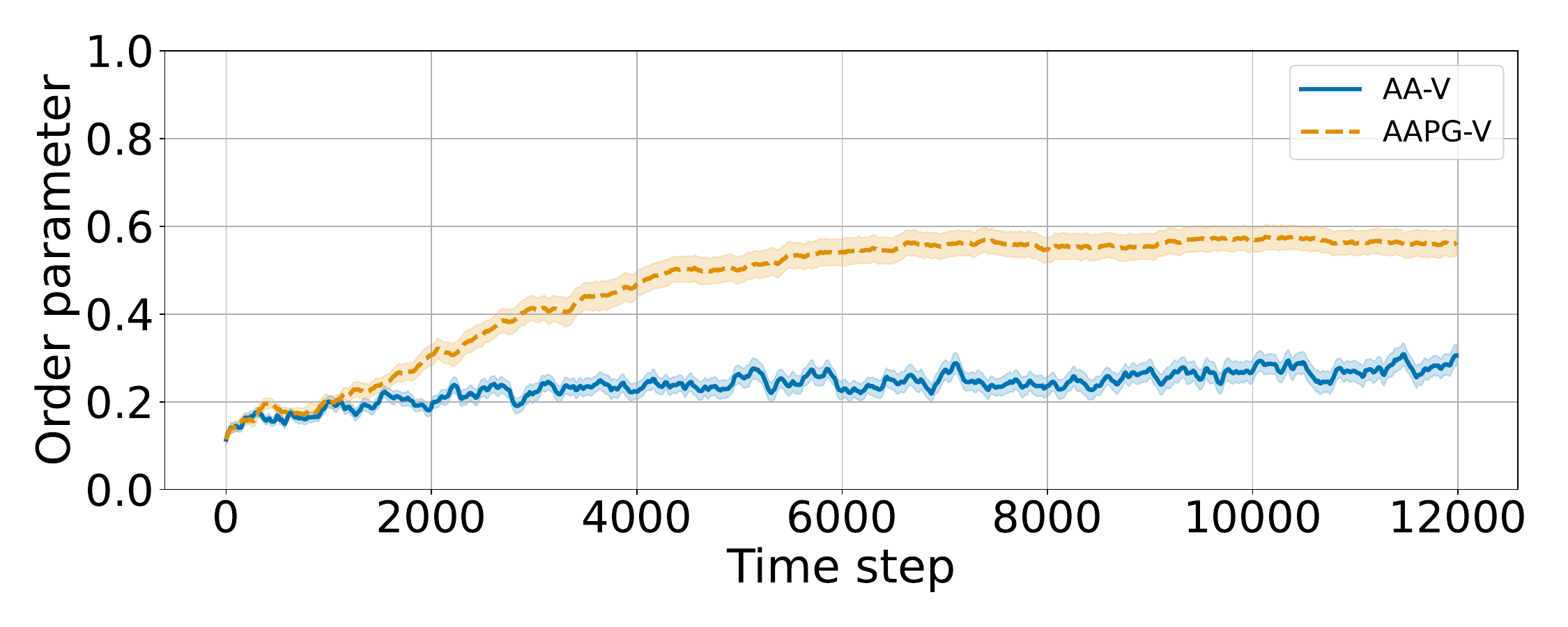}
        \caption{60 robots, 20\% faulty.}
    \end{subfigure}

    \caption{Summary comparison of motion order across swarm sizes (\emph{columns}: 25, 40, 60) and fault levels (\emph{rows}: 0\%, 10\%, 20\%). Each panel contrasts AA-V and AAPG-V.}
    \label{fig:compare_results}
\end{figure*}

\Cref{fig:compare_results} shows the evolution of the order parameter for swarms of 25, 40, and 60 robots (columns), each with 0\%, 10\%, and 20\% faulty robots (rows).
In all cases, AAPG-V either maintains or improves improves robustness to failures and outperforms the AA-V baseline. The improvements are more pronounced as the number of faulty robots increases.

\paragraph{Collective Motion Velocity}
P\&G inevitably slows down the swarm, since a fraction of the robots are paused at any given time. In fault-free conditions, this may reduce the effective collective speed compared to the standard AA-V model.  We therefore measure the speed of the swarm in different settings.

\begin{figure*}[htbp]
	\centering
	\begin{subfigure}{0.32\textwidth}
		\centering      \includegraphics[width=\linewidth]{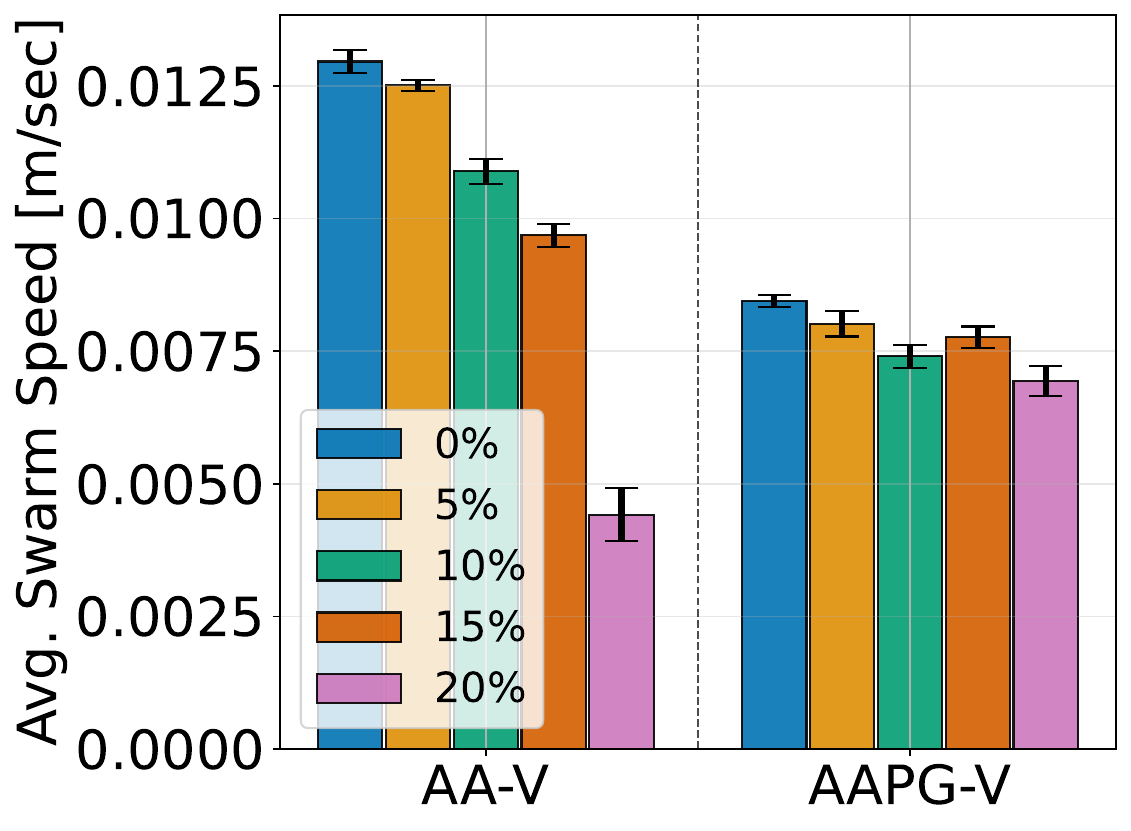}
		\caption{25 robots.}
		\label{fig:swarm_vel_25}
	\end{subfigure}
	\hfill
	\begin{subfigure}{0.32\textwidth}
		\centering
		\includegraphics[width=\linewidth]{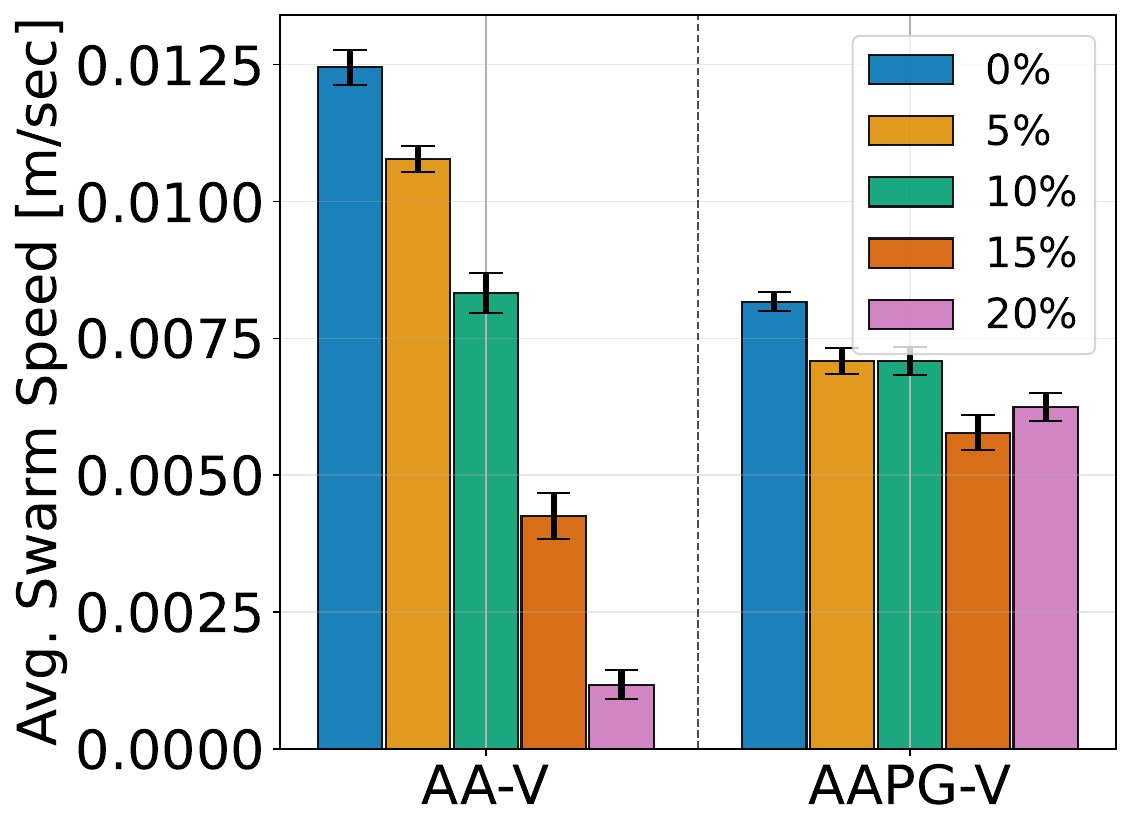}
		\caption{40 robots.}
		\label{fig:swarm_vel_40}
	\end{subfigure}
	\hfill
\begin{subfigure}{0.32\textwidth}
		\centering
		\includegraphics[width=\linewidth]{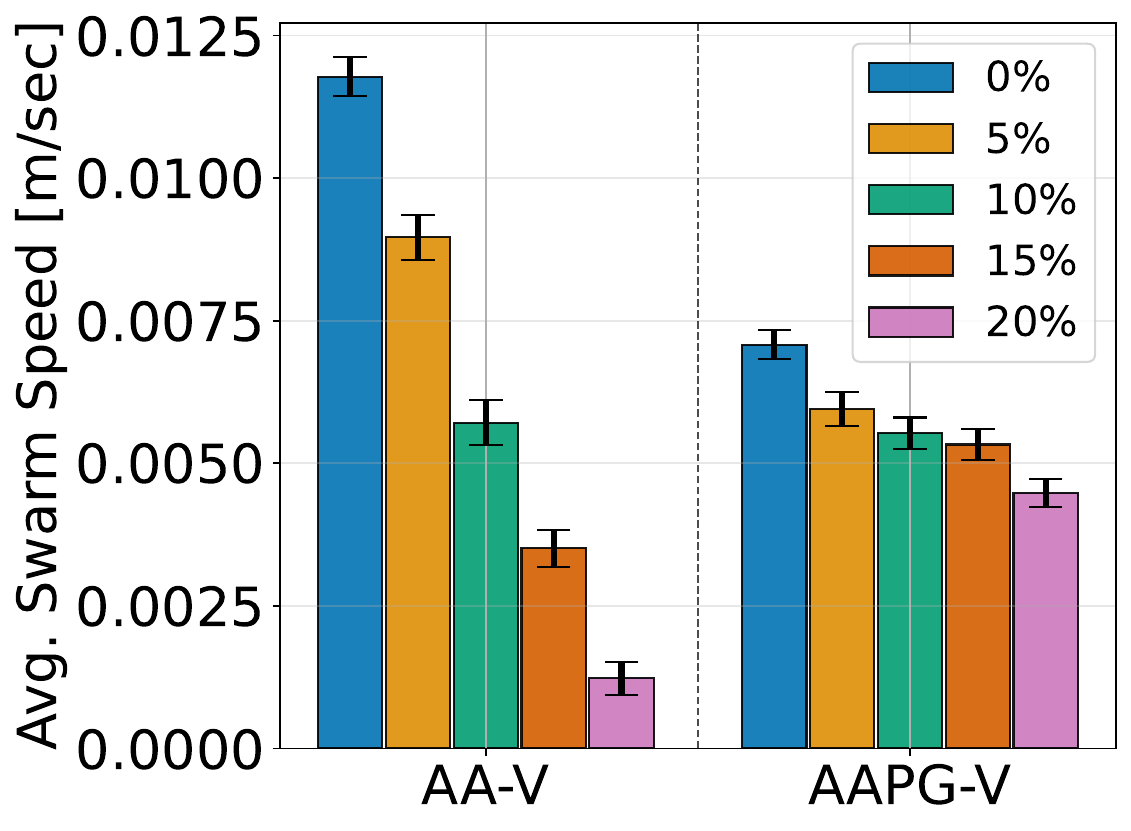}
		\caption{60 robots.}
		\label{fig:swarm_vel_60}
	\end{subfigure}

	\caption{Average speed of AA-V and AAPG-V across varying faulty proportions for different swarm sizes.}
	\label{fig:swarm_vel}
\end{figure*}

\cref{fig:swarm_vel} shows the swarm speed for different swarm sizes and faulty robot percentages. In each sub-figure, the left group of bars presents the average speed achieved with AA-V. The right group shows the speeds with AAPG-V. As the proportion of failed agents increases, the speed of the AA-V swarm deteriorates sharply, as does the order (see~\cref{fig:multi_faulty_robots}).
In contrast, the AAPG-V model generates a slower speed at a low percentage of faulty robots, but robustly maintains this speed when more faulty robots are present.

\section{Discussion}
\label{ch:Discussion}

Several topics arise from the experiments and are discussed in this section. First, many models of collective motion do not only rely on the range and bearing to neighbors, but also on observation of the neighbor's alignment (orientation relative to the heading of the focal robot). \Cref{sec:aaa_model} discusses the use of P\&G locomotion in such models. Second, the experiments have tested the P\&G mechanism in the presence of catastrophic failures, where the faulty robots become  stationary. \Cref{sec:slowed_down_robots} discusses less extreme failures.  Finally, \cref{sec:discuss-others} discusses alternative occlusion-handling methods.

\subsection{P\&G in Alignment Models}
\label{sec:aaa_model}
The P\&G method can potentially be extended to a wider range of collective motion models, other than those relying on distance estimation (from which velocities are derived).
To demonstrate this, we show the effectiveness of the approach when used with an \emph{Avoid-Align-Attract} (AAA) model, extending the AA model described above.
Such a model was introduced by \citet{ferrante12}.
It extends the computed AA motion vector  $\vec{F^t}$ (\cref{eq:avoid-attract}) by adding an alignment vector (\cref{eq:avoid-align-attract}):
\begin{equation}
	\vec{F_{align}^t} = \vec{F^t} + K_3 \cdot \vec{f_{align}^t}.
	\label{eq:avoid-align-attract}
\end{equation}
Here, $\vec{f_{align}^t}$ (\cref{eq:align}) represents the Vicsek alignment force~\citep{vicsek95}, resulting from aggregating the vectors $\vec{v_{ij}}$. Each vector  $\vec{v_{ij}}$ is the velocity difference between the focal robot $i$ and neighbor $j$. Normalizing $\vec{v_{ij}}$ gives the heading difference, and averaging over all neighbors produces the alignment force $\vec{f_{align}^t}$.
\begin{equation}
	\vec{f_{align}^t}= \sum_{j=1}^{n_i^t} \frac{\vec{v_{ij}}}{\| \vec{v_{ij}} \|}.
	\label{eq:align}
\end{equation}
Previous work has discussed how such vectors $\vec{v_{ij}}$ can be estimated from vision~\citep{moshtag09,krongauz24}, and so we do not discuss it here.
We refer to the resulting model as AAA-V (using $K_3 = 1.5$, determined empirically).

\begin{figure}[htbp]
	\centering
\includegraphics[width=0.99\columnwidth]{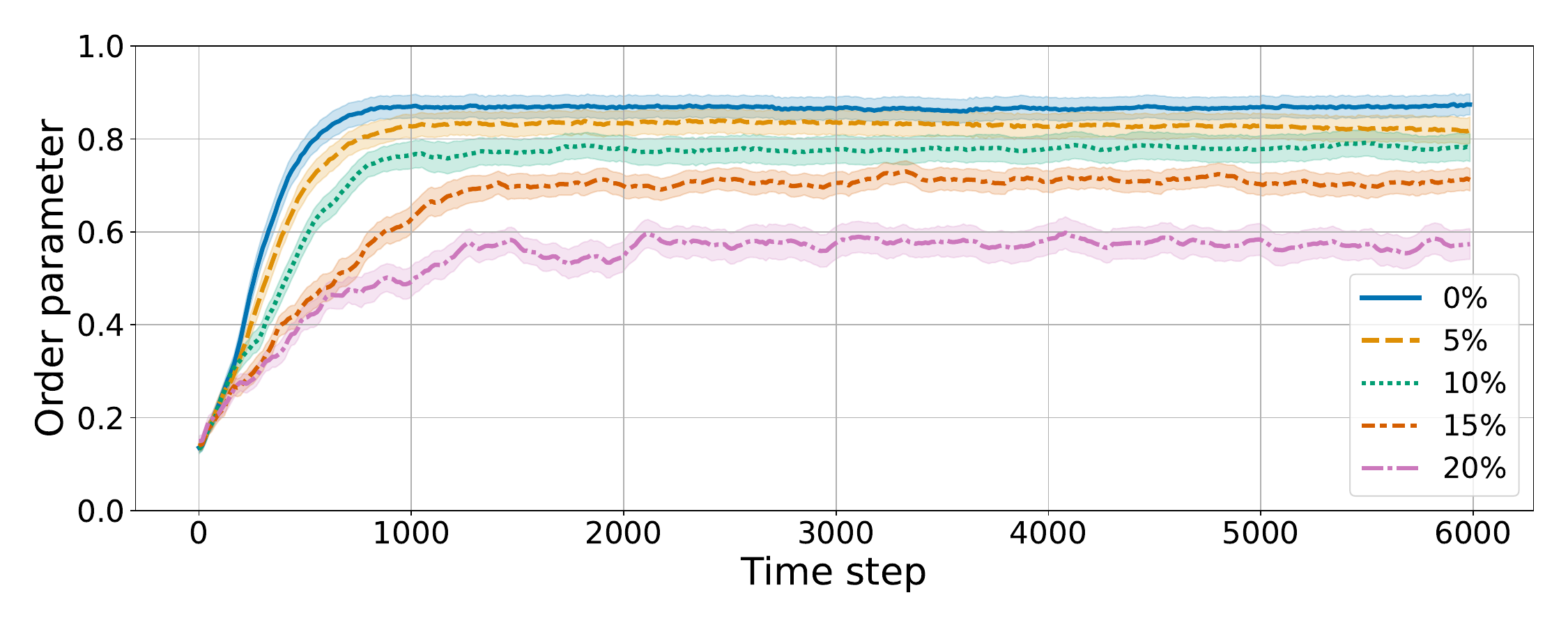}
		\caption{AAA-V with different percentages of faulty robots (40 robots total).}
		\label{fig:aaa_df_faulty}
\end{figure}

\begin{figure}[htbp]
	\centering

\includegraphics[width=0.99\columnwidth]{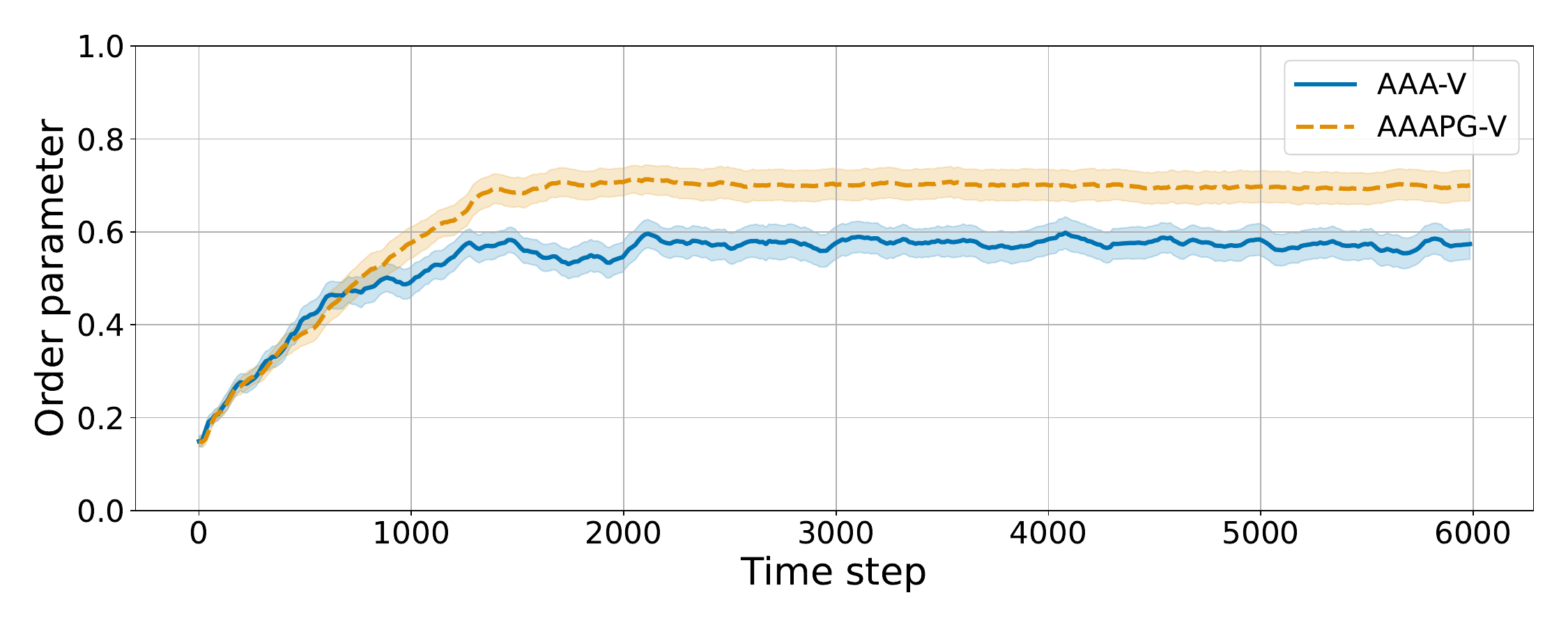}
	\caption{AAA-V vs AAAPG-V with 20\% faulty robots, out of 40 robots total. AAAPG-V clearly better.}
		\label{fig:aaa_10_faulty}

\end{figure}

\Cref{fig:aaa_df_faulty,fig:aaa_10_faulty} show two key results experimenting with AAA-V model and its extension using P\&G locomotion (AAAPG-V).
First,~\cref{fig:aaa_df_faulty} shows the evolving order of a 40-robot swarm using AAA-V, with different percentages of faulty robots.

The degradation in order is evident: as more robots are faulty, the order achieved is clearly lower.
Then, in~\cref{fig:aaa_10_faulty} we contrast the AAA-V and the AAAPG-V models, in the case of 20\% faulty robots.  The P\&G mechanism employs the same duration settings as in AAPG-V and incorporates the Avoid Half Time method, resulting in a markedly higher level of emergent order.

\subsection{Slowed-Down Robots}
\label{sec:slowed_down_robots}
Not all faults result in complete immobility. Some may cause robots to slow down without stopping entirely.
\Cref{def:failed_robot} covers such failures, relying on the thresholds $U_{min}$ and $\Theta_{min}$ to be set appropriately.

\begin{figure}
	\centering
	\includegraphics[width=\linewidth]{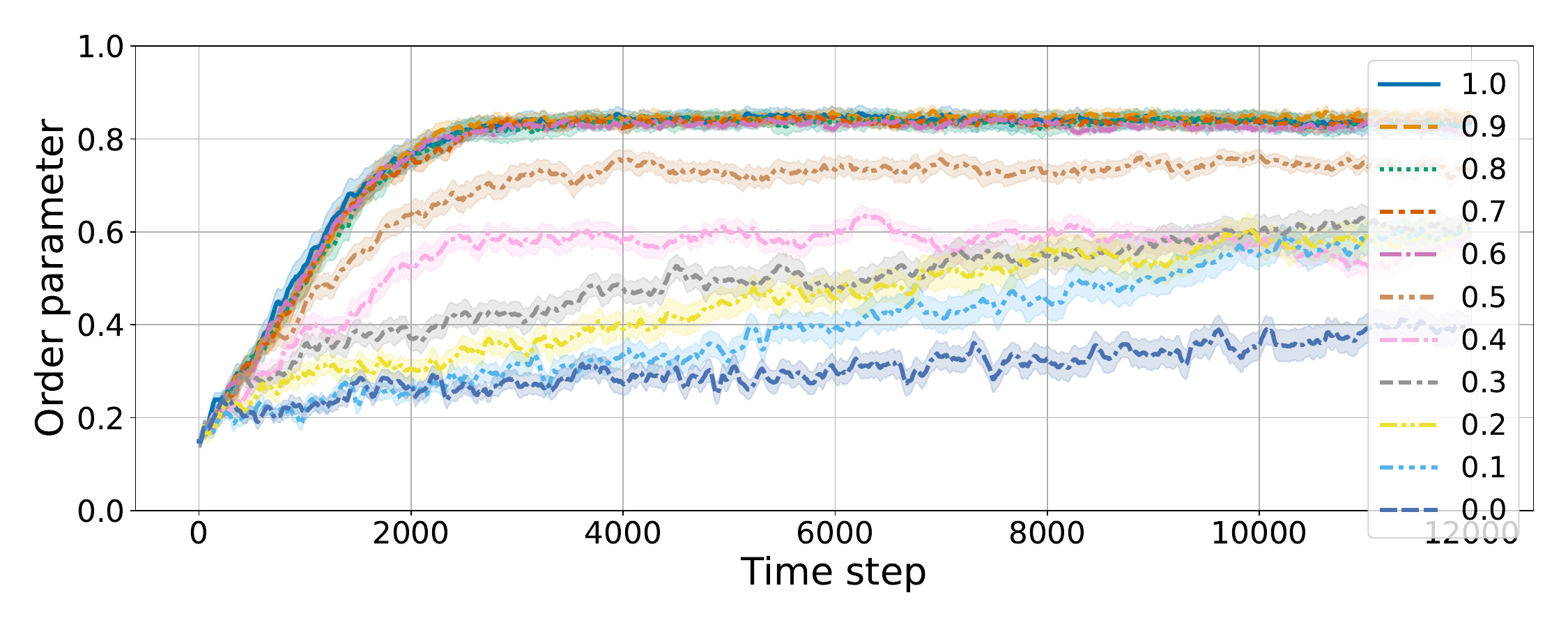}
	\caption{AA-V:  Order in swarms, where 20\% are slowed-down by different factors.}

	\label{fig:slow_sf}
\end{figure}

To empirically demonstrate this, we experimented with faulty robots that are slowed down. Their linear and angular speeds are multiplied by a slowdown factor ($S_f$). \Cref{fig:slow_sf} shows the effect of 20\% such faulty robots, when $S_f$ is varied from 1 (no slowdown) to 0 (complete stop).

\begin{figure}
	\centering
	\includegraphics[width=\linewidth]{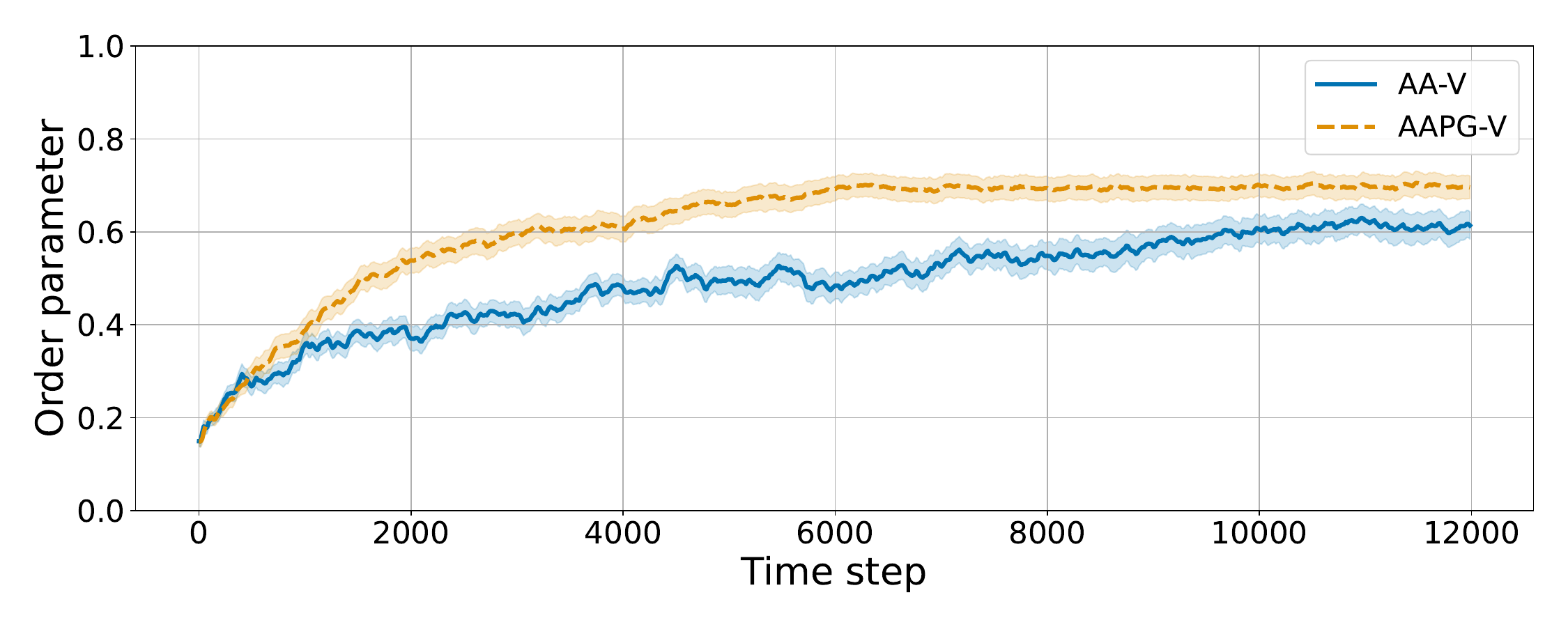}

	\caption{AA-V and AAPG-V with 20\% faulty robots (slowed, $S_f = 0.3$). }
\label{fig:slow_pg_methods}
\end{figure}

The objective in the following experiment is to show that the P\&G mechanism can be effective in detecting and avoiding faulty robots.
We set $S_f=0.3$, and empirically tuned the thresholds: the linear speed threshold is $U_{min}=0.0125\ms$, and the angular width change threshold is $\Theta_{min}=0.4$ (40\% change).
\Cref{fig:slow_pg_methods} compares the AA-V and AAPG-V models in a swarm of 40 agents with 20\% slowed down to $S_f = 0.3$, using these thresholds.
AAPG-V converges faster and reaches a higher order.

\subsection{Occlusions: A Delicate Issue}
\label{sec:discuss-others}
We are certainly not the first to discuss the role of occlusions in the perceptual processes that lead to neighbor detection and selection.  We have made a fairly common assumption that a neighbor can be visually recognized when visible. This leads to a spectrum of neighbor recognition capabilities: from OMID (recognize only when wholly visible) through CENTER (approximately: recognize when sufficiently visible), to COMPLID (recognize if any part is visible).  While these differ in how they treat partially visible neighbors, they share the minimal requirement that wholly visible neighbors are recognized as such.  This is not a trivial assumption.

\citet{ferrante12} recognize the importance of addressing occlusions in a manner that considers computational requirements. They report finding a delicate relationship between the sensing range ($r_{sense}$) and the desired inter-robot distance ($r_d$).
They note that at steady state, the most important interactions occur with respect to the first (closest) \emph{Voronoi neighbors}, which can be determined by Voronoi tessellation of the sensed area after neighbor recognition has been carried out as above (utilizing X-RAY model). Other neighbors contribute only marginally to the computed motion vector.

To avoid carrying out the neighbor recognition (and the tessellation) processes, \citet{ferrante12} propose a shortcut: given a target $r_d$, limit the sensing range to $r_{sense} = 1.8 \cdot 2^{\frac{1}{6}} \cdot r_d$. Indeed, for $r_d=0.10$, this sets the sensing range at 0.202, not far from our own set sensing range of 0.19.  For  $r_d=0.11$, the computed sensing range would be $0.22$.

However, setting an arbitrary $r_{sense}$ may not be possible: limits on effective $r_{sense}$ vary with sensor resolution, visual features used for recognition or registration, and even lighting conditions.  It may be more desirable, therefore, to solve for $r_d$, given a set $r_{sense}$.
However, this shortcut is not robust for such an inversion of purpose. For example, given a sensing range of 0.19, the computed $r_d$ would be 0.094. For this value, the X-RAY model would achieve a very low order.

This demonstrates that the values of $r_{sense}$ and $r_d$ should be set with great caution. Neither can be set arbitrarily.
The value of $r_d$ is constrained by stopping distances, robot velocity, and kinematic requirements.
Similarly, the value of $r_{sense}$ is constrained by the visual sensor capabilities---and the visual perception computations that utilize it.
Their relationship to the emerging order depends on many factors.

We therefore point to a large open area for future research. We need to consider methods for detecting neighbors in the presence of occlusions, and their possible interaction with the capabilities of the vision sensor, and the parameters of the collective motion model. These form a complex, tightly-coupled system of perceptual devices and processes in service of collective vision.

 \section{Summary}
\label{ch:Summary}
This paper presents two models for generating collective motion, using a visual sensor.  The AA-V model (introduced in~\cref{ch:pereception_challenges}) discusses the role of occlusions (and associated processes for recognizing neighbors), and introduces a neighbor range estimator that corrects for errors caused when perceiving elongated (rectangular box-like) robots, by considering the horizontal size of the neighbor (\cref{eq:aav}).  The AAPG-V model (introduced in~\cref{ch:problem_analysis}) discusses an approach allowing detection and handling of robots that are faulty---too slow to keep up---so that they do not disrupt the ordered motion of the swarm. The approach is based on asynchronous intermittent locomotion: when a robot pauses moving, it is able to detect a lack of sufficient movement in others. This is infeasible when the robot is moving.
The two models are examined empirically in extensive physics-based simulations. Movies demonstrating the results are available at \url{https://www.cs.biu.ac.il/~galk/research/swarms/}.

Much remains for future work. The discussion~\cref{ch:Discussion} raises a number of areas for further exploration, chief among them is a need for a comprehensive investigation of the delicate relation between occlusions, model parameters, perceptual processes, and visual sensor limits. The experimental data reveals that the interval parameters $\underrightarrow{G},\underrightarrow{P}$ impact the order and misclassification rate.
However, we do not yet have a theory relating these parameters to the success of the collective motion in terms of order or speed.

The techniques presented in this paper are inspired by natural swarm phenomena.
These appear to offer additional inspirations.
For example, individual fish flocking in schools exhibit an analogous phenomenon to intermittent motion, called \emph{burst and coast} (or \emph{burst and glide}), where a fish swims for a short duration (the burst), and then allows the momentum and currents to carry it for a short duration (coast/glide)~\citep{calovi18}.  During coasting, the visual field of the fish does change, and so using this period to detect failing peers does not appear to be doable in the way described above.
We believe that the actual computation carried out during motion pauses (including coasting) is more involved, and perhaps provides more information for the use of the individual swarm member.
We hope that future research will permit us to learn more from these natural swarms.

\appendix

\section{The Avoid--Attract Model}
\label{app:aa}

This appendix expands on the brief description in Section~\ref{sec:aa_model}, providing the full mathematical specification of the Avoid--Attract (AA) model, including motion computation, control laws, and implementation parameters.
The formal specification of the \emph{sensing}, \emph{neighbor detection}, \emph{neighbor selection} appears in~\cref{sec:baseline}.
For the \emph{motion computation} stage, we use \cref{eq:avoid-attract,eq:rjvector}, with $K_f=1$.  \Cref{tab:aa_params_appendix} lists the key notations.

The final stage (\emph{motion control}) uses the unmodified AA model equations.
The flocking vector $\vec{F_i^t}$ is mapped into linear and angular velocity commands via magnitude-dependent motion control (MDMC)~\citep{ferrante12}. Let \( (f_x, f_y) \) be the components of \( \vec{F^t} \) in robot \( i \)’s local frame.  Then
\begin{equation}
	u = K_{1} f_x + U_{\text{forward}},
	\qquad
	\omega = K_{2} f_y ,
	\label{eq:mdmc_appendix}
\end{equation}
where \( u \) is the linear velocity, \( \omega \) the angular velocity, \( K_1,K_2 \) are gain constants, and \( U_{\text{forward}} \) is half the maximum speed \( U_{\text{max}} \). Physical constraints are enforced:
\[
u \in [0, U_{\text{max}}],
\qquad
\omega \in [-\omega_{\text{lim}}, \omega_{\text{lim}}],
\]
where \( \omega_{\text{lim}} \) is the maximum angular velocity.

Finally, \( u \) and \( \omega \) are converted into right (\( W_r \)) and left (\( W_l \)) wheel velocities for differential drive:
\begin{equation}
	W_r = u - \tfrac{\omega l_w}{2},
	\qquad
	W_l = u + \tfrac{\omega l_w}{2},
	\label{eq:diff_drive_appendix}
\end{equation}
where \( l_w \) is the wheelbase.

\begin{table*}[htpb]
\centering
\begin{tabular}{lcl}
\toprule
\textbf{Symbol} & \textbf{Typical Value} & \textbf{Description (units)} \\
\midrule
$\mathcal{A}$        & --  & Set of all robots in the swarm \\
$\mathcal{D}_i^t$    & --           & Robots detected by robot $i$ at time $t$  \\
$\mathcal{N}_i^t$    & --           & Robots selected as \emph{nominal} by robot $i$ at time $t$  \\
$\mathcal{S}_i^t$    & --           & Robots selected as \emph{faulty} by robot $i$ at time $t$  \\
\midrule
$\vec{F_i^t}$    & --           & Motion vector of robot $i$ at time $t$ (polar) \\
$\vec{f_{ij}^t}$    & --           & AA vector of robot $i$ with neighbor $j$ at time $t$ (polar) \\
$\vec{f_{align}^t}$    & --           & Alignment vector of robot $i$ with neighbor $j$ at time $t$ (polar) \\
$r_{\text{sense}}$ & $0.19$ & Maximum neighbor detection distance (m) \\
$\Ct{}$  & --           & Neighbor $j$'s center in $i$'s polar system (m, rad) \\
$r_{ij}, r_C$      &--         & (Estimated) distance between robot $i$ and $j$ (m) \\
$\beta_{ij}, \beta_C$  & --           & Estimated bearing to neighbor center (rad) \\
$\angle\beta_{ij}$      & --         & Bearing angle between robot $i$ and $j$ (rad) \\
$r_d$      & $0.10$         & Preferred inter-robot spacing (m) \\
$K_f$      & $1.0$         & Avoid--Attract force gain  \\
$K_1$      & $2.5$          & MDMC linear gain  \\
$K_2$      & $0.06$         & MDMC angular gain  \\
$K_3$      & $1.5$         & Align gain in AAA model  \\
$u$         & --         & Linear velocity of the robot (m/s) \\
$\omega$      & --         & Angular velocity of the robot (rad/s) \\
$\hat{v}$      & --         & Unit vector speed (m,rad) \\
$U_{\text{max}}$     & $0.035$        & Maximum linear velocity (m/s) \\
$U_{\text{forward}}$ & $0.0175$       & Forward bias velocity (m/s) \\
$\omega_{\text{lim}}$& $0.5236$         & Maximum angular velocity (rad/s) \\
\midrule
$G_i^t$    & --           & Motion vector of robot $i$ at time $t$ with fault handling(polar) \\
$\vec{s_{ij}^t}$    & --           & AA vector of robot $i$ with faulty neighbor $j$ at time $t$ (polar) \\
$l$       & $0.05$      & Robot length (m) \\
$w$       & $0.02$      & Robot width (m) \\
$v$       & $0.027$     & Robot vertical size (height) (m) \\
$l_w$      &$ 0.018$        & Wheelbase (m) \\
$L,\ R$    & --           & Outermost visible vertical edges of the neighbor  \\
$\Lt{},\ \Rt{}$    & --           & Vectors to outermost visible vertical edges of the neighbor  \\
$\beta_L,\ \beta_R,$ & --  & Bearing angles to $L$, $R$ (rad) \\
$r_L,\ r_R$& --           & Horizontal ranges to $L$ and $R$ (m) \\
$\theta$ & --           & Horizontal subtended angle: $\beta_R-\beta_L$ (rad) \\
$\alpha_L, \alpha_R$ & --           & Vertically subtended angles (rad) \\
$H$      & --           & Metric length of visible horizontal edge (m) \\
$\varepsilon$ & 0        & Tolerance for matching $H$ to $l$ or $w$ (m) \\
$\vec{\delta_U}$ & -- & Instantaneous position change of a neighbor (m, rad) \\
$\delta_U$, $\delta_\theta$ & -- & Linear and rotational velocities of a neighbor (m/s), (rad/s) \\
 $U_{min}, \Theta_{min}$& -- & Linear and rotational thresholds (m/s), (\%rad/s)\\
$\underrightarrow{P} ,\ \underrightarrow{G}$& $[6,7), \ [11,20)$ & Pause and go range durations (ticks)\\
$\mathbb{H}$ & -- & Heaviside function (1 when positive, 0 otherwise). \\
\bottomrule
\end{tabular}
\caption{Nomenclature: Parameters and symbols used in the Avoid--Attract (AA) model and the vision-based neighbor detection. Typical values apply to our simulations/experiments where specified.}
\label{tab:aa_params_appendix}
\end{table*}

\section{P\&G Sensitivity}
\label{app:pgsensitive}
We provide the detailed results from extensive experiments with 24 variations of the pause interval parameter $\underrightarrow{P}$, combined with 6 variations of the go interval parameter $\underrightarrow{G}$.  There are 144 combinations; each was tried 50 times, with different seeds, so as to establish reasonable confidence in the value.
Tables~\cref{tab:pg_fp,tab:pg_o} present the results, where $\underrightarrow{P}$ variations are shown in rows, and $\underrightarrow{G}$ variations in columns.

\Cref{tab:pg_fp} shows the average number of misclassified neighbors (false positives) per focal robot, in these 144 combinations. All results are for a swarm of 40 robots with 0 faulty robots. This represents a worst-case bound on the number of misclassified neighbors.  Overall, the average number of false positives is between 1 (or just under it) to 2 (or just over it).
This represents between 10\%--20\% of neighbors, and generates a clear distortion of local information that cascades through the swarm.

\Cref{tab:pg_o} shows the corresponding mean order for the same runs, in the last 100 time steps (10 seconds) of the experiment (allowing plenty to time for the swarm to converge to ordered motion).
Once again, $\underrightarrow{P}$ is shown as rows, and $\underrightarrow{G}$ in columns. Each cell represents the mean over 50 trials. Although there is obvious variance in the different combinations,
generally the order is maintained between 0.7 and 0.85.

This represents between 10\%--20\% of neighbors, and generates a clear distortion of local information that cascades through the swarm.

\clearpage
\begin{table}[p]
\centering
\caption{Average false positive count per cycle (mean of 50 runs). Swarm size is 40 robots.}
\label{tab:pg_fp}
\begin{tabular}{lrrrrrr}
\toprule
{} & 10--12 & 11--14 & 11--18 & 11--20 & 11--25 & 20--35 \\
\midrule
6-7 & 1.972 & 1.881 & 1.731 & 1.699 & 1.506 & 1.138 \\
7-8 & 1.893 & 1.807 & 1.660 & 1.572 & 1.466 & 1.105 \\
8-9 & 1.783 & 1.714 & 1.585 & 1.546 & 1.398 & 1.057 \\
9-10 & 1.737 & 1.678 & 1.551 & 1.488 & 1.369 & 1.036 \\
10-11 & 1.690 & 1.600 & 1.505 & 1.451 & 1.307 & 1.019 \\
11-12 & 1.648 & 1.529 & 1.464 & 1.401 & 1.284 & 0.985 \\
12-13 & 1.597 & 1.501 & 1.396 & 1.368 & 1.254 & 0.989 \\
13-14 & 1.562 & 1.463 & 1.378 & 1.319 & 1.218 & 0.965 \\
14-15 & 1.508 & 1.417 & 1.332 & 1.295 & 1.201 & 0.941 \\
\midrule
5-7 & 2.086 & 1.921 & 1.815 & 1.728 & 1.568 & 1.164 \\
6-8 & 1.962 & 1.835 & 1.717 & 1.631 & 1.479 & 1.120 \\
7-9 & 1.887 & 1.763 & 1.623 & 1.558 & 1.414 & 1.081 \\
8-10 & 1.770 & 1.681 & 1.576 & 1.514 & 1.391 & 1.047 \\
9-11 & 1.721 & 1.640 & 1.500 & 1.457 & 1.351 & 1.022 \\
10-12 & 1.646 & 1.553 & 1.487 & 1.427 & 1.305 & 1.008 \\
11-13 & 1.579 & 1.515 & 1.445 & 1.383 & 1.277 & 0.992 \\
12-14 & 1.547 & 1.473 & 1.385 & 1.327 & 1.252 & 0.965 \\
13-15 & 1.514 & 1.434 & 1.349 & 1.326 & 1.210 & 0.957 \\
\midrule
5-8 & 2.023 & 1.873 & 1.750 & 1.668 & 1.536 & 1.158 \\
6-9 & 1.922 & 1.802 & 1.681 & 1.587 & 1.487 & 1.101 \\
7-10 & 1.833 & 1.711 & 1.621 & 1.537 & 1.415 & 1.071 \\
8-11 & 1.759 & 1.648 & 1.548 & 1.476 & 1.362 & 1.036 \\
9-12 & 1.706 & 1.616 & 1.483 & 1.449 & 1.318 & 1.010 \\
10-13 & 1.659 & 1.555 & 1.453 & 1.403 & 1.290 & 0.995 \\
11-14 & 1.595 & 1.494 & 1.421 & 1.365 & 1.254 & 0.982 \\
12-15 & 1.532 & 1.464 & 1.368 & 1.338 & 1.234 & 0.954 \\
\midrule
5-9 & 1.954 & 1.859 & 1.716 & 1.630 & 1.478 & 1.108 \\
6-10 & 1.899 & 1.780 & 1.650 & 1.581 & 1.438 & 1.083 \\
7-11 & 1.818 & 1.708 & 1.573 & 1.503 & 1.378 & 1.046 \\
8-12 & 1.745 & 1.622 & 1.529 & 1.468 & 1.345 & 1.028 \\
9-13 & 1.682 & 1.567 & 1.470 & 1.431 & 1.305 & 1.010 \\
10-14 & 1.624 & 1.536 & 1.433 & 1.388 & 1.260 & 0.983 \\
11-15 & 1.562 & 1.482 & 1.398 & 1.344 & 1.243 & 0.978 \\
\midrule
5-10 & 1.938 & 1.852 & 1.685 & 1.625 & 1.474 & 1.106 \\
6-11 & 1.885 & 1.750 & 1.635 & 1.537 & 1.412 & 1.095 \\
7-12 & 1.795 & 1.673 & 1.544 & 1.500 & 1.375 & 1.041 \\
8-13 & 1.723 & 1.629 & 1.501 & 1.468 & 1.335 & 1.023 \\
9-14 & 1.662 & 1.574 & 1.455 & 1.395 & 1.288 & 1.007 \\
10-15 & 1.612 & 1.519 & 1.428 & 1.365 & 1.270 & 0.988 \\
\midrule
5-11 & 1.914 & 1.825 & 1.688 & 1.601 & 1.458 & 1.106 \\
6-12 & 1.861 & 1.723 & 1.624 & 1.550 & 1.409 & 1.067 \\
7-13 & 1.789 & 1.669 & 1.539 & 1.486 & 1.375 & 1.038 \\
8-14 & 1.722 & 1.621 & 1.519 & 1.455 & 1.324 & 1.016 \\
9-15 & 1.645 & 1.578 & 1.451 & 1.406 & 1.305 & 1.008 \\
\midrule
5-12 & 1.939 & 1.834 & 1.662 & 1.617 & 1.456 & 1.100 \\
6-13 & 1.833 & 1.731 & 1.614 & 1.550 & 1.428 & 1.074 \\
7-14 & 1.771 & 1.679 & 1.557 & 1.477 & 1.372 & 1.054 \\
8-15 & 1.728 & 1.607 & 1.504 & 1.456 & 1.313 & 1.030 \\
\midrule
5-13 & 1.949 & 1.837 & 1.679 & 1.613 & 1.461 & 1.114 \\
6-14 & 1.852 & 1.704 & 1.607 & 1.567 & 1.419 & 1.085 \\
7-15 & 1.807 & 1.660 & 1.579 & 1.491 & 1.383 & 1.061 \\
\midrule
5-14 & 1.953 & 1.819 & 1.700 & 1.623 & 1.482 & 1.142 \\
6-15 & 1.866 & 1.747 & 1.642 & 1.564 & 1.432 & 1.101 \\
\midrule
5-15 & 1.939 & 1.834 & 1.697 & 1.628 & 1.477 & 1.151 \\
\bottomrule
\end{tabular}
\end{table}

\begin{table}[p]
\centering
\caption{Average order achieved in the last 100 time steps (mean of 50 runs). }
\label{tab:pg_o}
\begin{tabular}{lrrrrrr}
\toprule
{} & 10--12 & 11--14 & 11--18 & 11--20 & 11--25 & 20--35 \\
\midrule
6-7 & 0.809 & 0.808 & 0.824 & 0.847 & 0.813 & 0.783 \\
7-8 & 0.797 & 0.824 & 0.793 & 0.787 & 0.840 & 0.799 \\
8-9 & 0.773 & 0.794 & 0.793 & 0.821 & 0.796 & 0.780 \\
9-10 & 0.801 & 0.845 & 0.807 & 0.810 & 0.802 & 0.802 \\
10-11 & 0.822 & 0.824 & 0.817 & 0.827 & 0.767 & 0.778 \\
11-12 & 0.847 & 0.773 & 0.805 & 0.792 & 0.770 & 0.717 \\
12-13 & 0.813 & 0.794 & 0.744 & 0.776 & 0.746 & 0.744 \\
13-14 & 0.795 & 0.792 & 0.762 & 0.776 & 0.729 & 0.700 \\
14-15 & 0.767 & 0.773 & 0.768 & 0.747 & 0.682 & 0.632 \\
\midrule
5-7 & 0.851 & 0.822 & 0.858 & 0.849 & 0.838 & 0.812 \\
6-8 & 0.825 & 0.821 & 0.814 & 0.832 & 0.815 & 0.812 \\
7-9 & 0.832 & 0.823 & 0.799 & 0.778 & 0.780 & 0.814 \\
8-10 & 0.804 & 0.812 & 0.818 & 0.811 & 0.807 & 0.809 \\
9-11 & 0.820 & 0.795 & 0.794 & 0.782 & 0.802 & 0.773 \\
10-12 & 0.800 & 0.793 & 0.814 & 0.818 & 0.784 & 0.762 \\
11-13 & 0.779 & 0.788 & 0.813 & 0.784 & 0.733 & 0.732 \\
12-14 & 0.806 & 0.773 & 0.782 & 0.764 & 0.746 & 0.688 \\
13-15 & 0.805 & 0.802 & 0.762 & 0.780 & 0.710 & 0.629 \\
\midrule
5-8 & 0.815 & 0.789 & 0.837 & 0.819 & 0.848 & 0.838 \\
6-9 & 0.792 & 0.809 & 0.817 & 0.827 & 0.832 & 0.802 \\
7-10 & 0.795 & 0.797 & 0.819 & 0.807 & 0.822 & 0.785 \\
8-11 & 0.783 & 0.777 & 0.804 & 0.788 & 0.815 & 0.769 \\
9-12 & 0.849 & 0.832 & 0.798 & 0.802 & 0.773 & 0.772 \\
10-13 & 0.826 & 0.806 & 0.782 & 0.792 & 0.765 & 0.750 \\
11-14 & 0.827 & 0.799 & 0.798 & 0.787 & 0.760 & 0.727 \\
12-15 & 0.796 & 0.803 & 0.792 & 0.766 & 0.728 & 0.668 \\
\midrule
5-9 & 0.788 & 0.824 & 0.818 & 0.814 & 0.782 & 0.790 \\
6-10 & 0.795 & 0.816 & 0.819 & 0.813 & 0.823 & 0.784 \\
7-11 & 0.823 & 0.817 & 0.792 & 0.786 & 0.773 & 0.753 \\
8-12 & 0.811 & 0.808 & 0.823 & 0.804 & 0.792 & 0.786 \\
9-13 & 0.832 & 0.799 & 0.808 & 0.810 & 0.782 & 0.752 \\
10-14 & 0.812 & 0.781 & 0.804 & 0.794 & 0.743 & 0.730 \\
11-15 & 0.819 & 0.779 & 0.799 & 0.755 & 0.741 & 0.726 \\
\midrule
5-10 & 0.798 & 0.834 & 0.805 & 0.825 & 0.808 & 0.808 \\
6-11 & 0.831 & 0.824 & 0.834 & 0.795 & 0.797 & 0.822 \\
7-12 & 0.801 & 0.806 & 0.750 & 0.796 & 0.799 & 0.767 \\
8-13 & 0.810 & 0.826 & 0.809 & 0.802 & 0.784 & 0.746 \\
9-14 & 0.795 & 0.787 & 0.769 & 0.781 & 0.781 & 0.743 \\
10-15 & 0.833 & 0.792 & 0.781 & 0.773 & 0.763 & 0.720 \\
\midrule
5-11 & 0.788 & 0.845 & 0.823 & 0.798 & 0.792 & 0.770 \\
6-12 & 0.818 & 0.778 & 0.836 & 0.824 & 0.796 & 0.771 \\
7-13 & 0.801 & 0.793 & 0.782 & 0.793 & 0.773 & 0.734 \\
8-14 & 0.802 & 0.802 & 0.822 & 0.812 & 0.778 & 0.725 \\
9-15 & 0.811 & 0.833 & 0.775 & 0.772 & 0.781 & 0.694 \\
\midrule
5-12 & 0.793 & 0.831 & 0.784 & 0.822 & 0.759 & 0.782 \\
6-13 & 0.759 & 0.791 & 0.785 & 0.798 & 0.800 & 0.771 \\
7-14 & 0.787 & 0.794 & 0.828 & 0.777 & 0.748 & 0.759 \\
8-15 & 0.813 & 0.790 & 0.781 & 0.785 & 0.747 & 0.708 \\
\midrule
5-13 & 0.813 & 0.802 & 0.809 & 0.789 & 0.789 & 0.736 \\
6-14 & 0.816 & 0.759 & 0.790 & 0.822 & 0.764 & 0.753 \\
7-15 & 0.837 & 0.777 & 0.821 & 0.774 & 0.759 & 0.739 \\
\midrule
5-14 & 0.838 & 0.811 & 0.810 & 0.797 & 0.778 & 0.782 \\
6-15 & 0.801 & 0.808 & 0.809 & 0.774 & 0.743 & 0.747 \\
\midrule
5-15 & 0.787 & 0.803 & 0.788 & 0.776 & 0.741 & 0.736 \\
\bottomrule
\end{tabular}
\end{table}

\clearpage

\end{document}